# Self-supervised DXA representations encode multi-system disease risk, biological aging and heritability

Gil Sasson[1], Zachary Levine[1], Smadar Shilo[1, 2, 3], Sarah Kohn[1], Guy Lutsker[1], Anastasia Godneva[1], Adam Gabet[1], David Krongauz[1], Adina Weinberger[1, 4], Yann LeCun[5, 6], Randall Balestriero[7] & Eran Segal[1, 8*]

**Author affiliations**

[1] Department of Computer Science and Applied Mathematics, Weizmann Institute of Science, Rehovot, Israel.

[2] Gray Faculty of Medical and Health Sciences, Tel Aviv University, Tel-Aviv, Israel.

[3] The Jesse Z and Sara Lea Shafer Institute for Endocrinology and Diabetes, National Center for Childhood Diabetes, Schneider Children's Medical Center of Israel, Petah Tikva, Israel.

[4] Department of Molecular Cell Biology, Weizmann Institute of Science, Rehovot, Israel.

[5] Courant Institute of Mathematical Sciences, New York University, New York, NY, USA.

[6] AMI Labs, Paris, France.

[7] Department of Computer Science, Brown University, Providence, RI, USA.

[8] Mohamed bin Zayed University of Artificial Intelligence, Abu Dhabi, UAE.

*** Corresponding author:**

Prof. Eran Segal, Department of Computer Science and Applied Mathematics, Weizmann Institute of Science, Rehovot, Israel, Tel: 972-8-934-3540, Fax: 972-8-934-4122, Email: eran.segal@weizmann.ac.il, ORCID: 0000-0002-6859-1164

## Abstract

Whole-body dual-energy X-ray absorptiometry (DXA) scans are routinely acquired to measure bone density and regional body composition, leaving their spatial structure largely unused. Here, we show that self-supervised learning (SSL) can convert raw DXA images into representations of systemic health. We introduce LeDXA, a vision model based on a joint-embedding predictive architecture (JEPA) that learns by predicting latent representations rather than reconstructing pixels. Trained from scratch on 11,540 unlabeled Human Phenotype Project scans, LeDXA was evaluated internally and on 47,400 external UK Biobank (UKBB) scans. It improved cross-cohort prediction of prevalent diseases and biomarkers beyond scanner-derived DXA measurements and DINOv3, a state-of-the-art general-purpose model, despite approximately 150,000-fold fewer training images and nearly 40-fold fewer parameters. Over a median 4.3-year UKBB follow-up, LeDXA improved incident disease prediction over tabular DXA measures, with the largest gains for hip and knee arthrosis and type 2 diabetes. For hip arthrosis, 66% of incident cases occurred in the highest-risk quartile versus 41% for tabular measures. Its representations predicted chronological age externally (r = 0.88; mean absolute error = 2.90 years), and the biological-age gap tracked broader disease burden and a 45% higher mortality hazard in the oldest-appearing quartile. The gap also decreased in women after starting hormone-replacement therapy, suggesting it may be modifiable. Genome-wide associations recovered mostly known body-composition and bone-density loci, and LeDXA embeddings were more heritable than DINOv3's. These findings reveal prognostic information in DXA images that conventional readouts discard, learnable with relatively little data and modest compute.

## Introduction

Progressive musculoskeletal decline and systemic metabolic dysfunction drive a growing global health burden across a spectrum of chronic conditions [1–3]. Dual-energy X-ray absorptiometry (DXA) occupies a unique position in this landscape: a low-radiation, low-cost imaging modality already acquired at population scale for osteoporosis screening [4], capturing both skeletal density and soft-tissue distribution in one screening session [5]. Beyond the areal bone mineral density (BMD), utilized in current clinical practice, DXA-derived measures of visceral adipose tissue (VAT), appendicular lean mass (ALM), and regional fat distribution have demonstrated prognostic value across cardiovascular disease, type 2 diabetes (T2D), obstructive sleep apnea, and all-cause mortality [6–9].

Yet these clinically useful measures represent only a compressed view of the DXA image. Their prognostic value suggests DXA captures systemic biological variation relevant to chronic disease, but current measurements access this variation through a small set of predefined summaries. BMD reduces skeletal information to a regional density average, while soft-tissue measures aggregate mass within broad anatomical compartments, discarding spatial variation in bone texture, fat deposition, and lean-tissue distribution across the body. The trabecular bone score (TBS) illustrates this: it extracts bone-texture information from the DXA image that standard BMD does not capture, showing the scan holds skeletal signal beyond regional density [10,11]. However, how much disease-relevant signal the raw scan contains beyond these predefined summaries remains unknown, and answering this requires learning representations directly from the scans rather than from their derived measurements.

So far, deep learning applied to DXA images has been mostly supervised; trained to predict specific outcomes like fracture risk or mortality [12,13]. This demonstrates the main limitation of supervised learning: every new target needs its own clinical labels, which are expensive to gather at a population scale and restrict the model to a single task [14]. Self-supervised learning (SSL) overcomes this limitation by learning transferable representations from unlabeled images. While SSL has led to powerful vision foundational models (VFMs) in retinal imaging [15], radiology [16], and pathology [17–19], it remains rarely used for DXA, predominantly focusing on narrow clinical endpoints rather than general-purpose representations [20].

Building a dedicated VFM for DXA is limited by dataset size. Standard DXA datasets are much smaller than the millions of images used to train foundation models in other fields, making it hard to train a new model from scratch. To bypass this, the standard workaround across the broader medical artificial intelligence (AI) field is to use transfer learning [21] from general-purpose models like DINOv2/v3 [22,23] or SAM [24], applying their frozen features to medical scans [25–27]. However, these models are tuned for the colors and edges of natural photos, not the physical density measurements of X-rays [28]. As a result, their frozen features focus on local, patch-level details [29,30]. This localized focus is poorly suited for studying chronic diseases, which typically cause body-wide, systemic changes rather than isolated abnormal regions.

Consequently, overcoming this gap requires training a domain-specific model. One framework for this is the joint-embedding predictive architecture (JEPA), an SSL method that learns by predicting target patterns in the latent space rather than reconstructing raw pixels. This encourages the model to focus on meaningful anatomical shapes rather than scanner noise, producing consistent results across natural images [31], video [32], and functional magnetic resonance imaging (fMRI) [33]. Building on this, the recent Latent-Euclidean JEPA (LeJEPA) formulation [34] introduces a highly stable training process, making it feasible to train robust models from scratch using standard-sized medical cohorts, avoiding the need for intricate engineering.

Here, we present LeDXA, a vision foundation model trained from scratch on whole-body DXA scans using the LeJEPA framework [34] (Fig. 1 and Fig. S1). Trained on 11,540 scans from the Human Phenotype Project (HPP) [35], a longitudinal study with extensive clinical and molecular profiling, LeDXA extracts clinically meaningful, whole-body representations of body composition. For downstream tasks, LeDXA outperforms standard tabular baselines as well as a state-of-the-art (SOTA) general model (DINOv3) trained on massive internet-scale datasets [23] , despite approximately 150,000-fold fewer training images and nearly 40-fold fewer parameters. It generalizes well to 47,400 external UK Biobank (UKBB) scans [36], maintaining diagnostic accuracy for prevalent disease. Over longitudinal follow-up, the model predicted incident conditions including T2D, arthrosis, and osteoporosis, while a LeDXA-derived biological age metric stratified all-cause mortality risk. Supporting the biological basis for these representations, a genome-wide association study (GWAS) on the UKBB data recovered established genetic loci for body morphology, along with loci associated with traits across multiple organ systems.

Together, these findings show that SSL can unlock clinically and biologically meaningful information from routine DXA scans beyond conventional densitometric features.

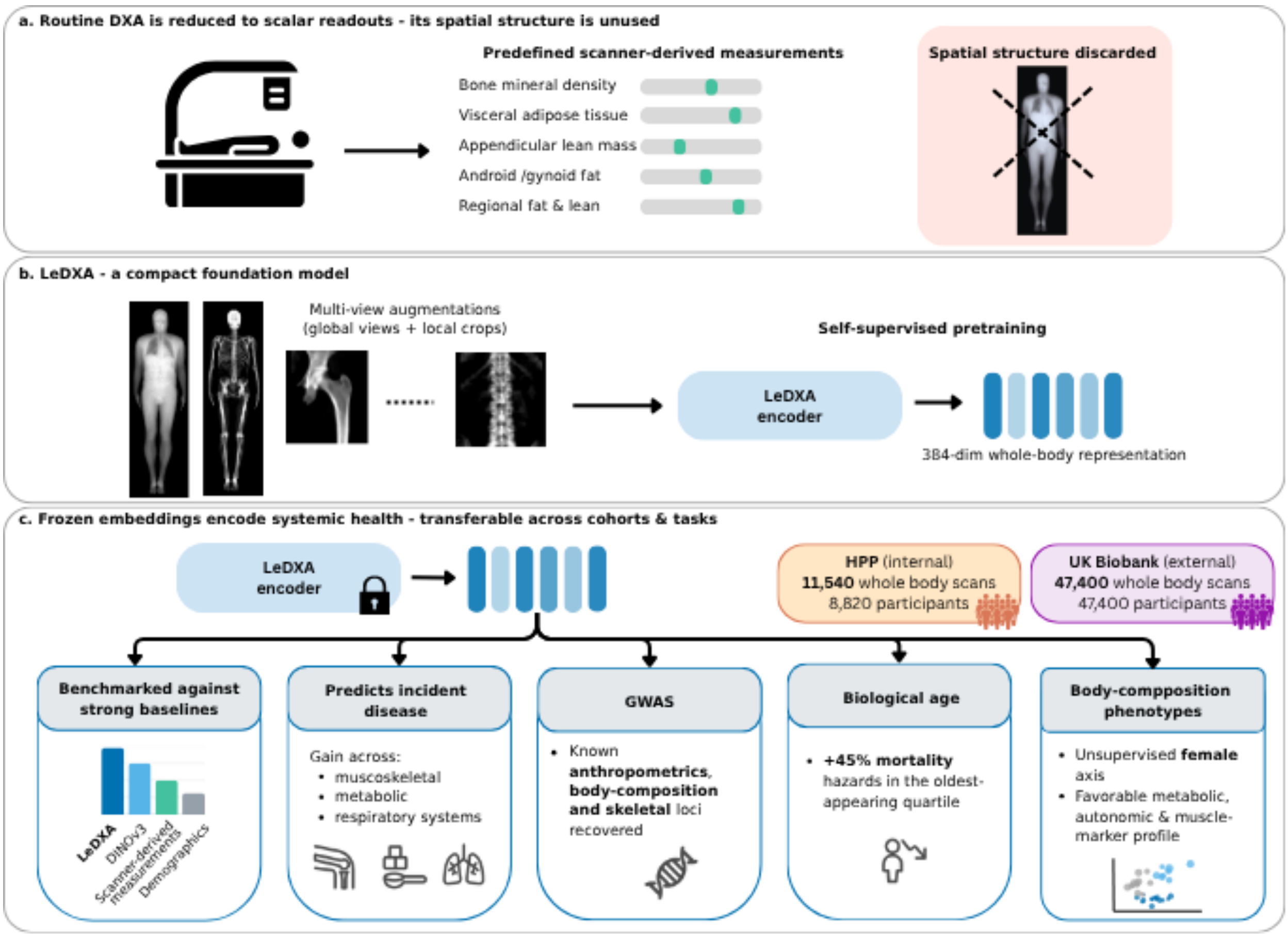

**Figure 1 | LeDXA converts DXA scans into representations that encode systemic health. a,** Routine DXA is reduced to a set of scalar readouts (bone mineral density, visceral adipose tissue, appendicular lean mass, android/gynoid fat, regional fat and lean), the spatial structure of the scan is left under-utilized. **b,** LeDXA is a compact self-supervised foundation model. Each whole-body scan is expanded into multi-view augmentations (2 global views and 8 local crops) and used to pretrain a ViT-Small/16 encoder, producing a 384-dimensional whole-body representation. **c,** The frozen embedding encodes systemic health and transfers across cohorts and tasks. Representations were learned on HPP (internal; 11,540 whole-body scans from 8,820 participants) and validated on UK Biobank (external; 47,400 whole-body scans from 47,400 participants). Across downstream tasks, LeDXA was benchmarked against strong baselines (DINOv3, scanner-derived measurements, and demographics), predicted incident musculoskeletal, metabolic, and respiratory disease, recovered known body-composition and bone loci from a genome-wide association study of its embeddings, stratified all-cause mortality (a 45% higher hazard in the oldest-appearing quartile), and recovered body-composition phenotypes including an unsupervised female axis linked to a favorable metabolic, autonomic, and muscle-marker profile.

## Results

### Study populations and DXA imaging

We pretrained LeDXA on DXA images from the HPP, a prospective deep-phenotyping cohort of generally healthy Israeli adults [35] (Methods). Each DXA scan uses two X-ray energies, producing paired bone- and soft-tissue images that separate bone mineral from soft tissue and partition soft tissue into lean and fat mass. Separate regional scans of the femur and lumbar spine (L1–L4) cover the standard sites for clinical bone-density assessment and osteoporosis diagnosis. We used 11,540 HPP scans from 8,820 participants for model development and internal evaluation. We validated the model externally in the UKBB [36], an older UK cohort spanning a wider range of health and disease, using 47,400 scans from the first imaging visit; no UKBB data were seen during pretraining. Baseline characteristics of both cohorts are summarized in Table S1. We applied no diagnosis or medication-based exclusions. Cohort selection and the resulting analytic samples are shown in Fig. S2; task-specific sample sizes are reported for each analysis (Supplementary information).

### LeDXA predicts multi-system disease and physiological markers

We first asked whether the LeDXA embedding retains the body-composition information that defines DXA's established clinical use. The embeddings recovered standard scanner-derived readouts with high fidelity, achieving Pearson $r = 0.964 \pm 0.005$ (mean ± SD) for fat mass, lean mass, regional adiposity, and visceral fat. DINOv3 recovered these measures comparably well ($r = 0.957 \pm 0.016$), confirming that both encoders capture basic compositional features (Fig. S3a). On the bone-density sites that define clinical diagnosis, LeDXA recovered femoral-neck, total-hip, and L1–L4 BMD significantly better than DINOv3 ($r = 0.882 \pm 0.026$ vs. $0.789 \pm 0.029$; corrected across sites by the Benjamini–Hochberg false discovery rate (FDR) procedure FDR-adjusted $P = 0.002$ at each site). For osteoporosis and osteopenia classification, we instead used features from the lumbar and femoral regional scans, matching the sites used clinically. With these regional features, LeDXA reached an area under the receiver operating characteristic curve (AUROC) of 0.87 for osteopenia and 0.91 for osteoporosis, approaching a classifier trained on scanner-derived BMD (0.90 and 0.95; Fig. S3b, c). LeDXA significantly exceeded DINOv3 in both (0.84 and 0.85; $P = 0.004$ and $P = 0.002$), indicating that domain-specific pretraining improves recovery of bone-density status from regional scans. Because these labels are self-reported and some participants may already be treated for bone loss, current scans may not reflect the original diagnosis.

To establish the model's capacity to capture continuous systemic health, we first benchmarked it on age and physiological markers prediction. For age, LeDXA achieved a mean absolute error (MAE) of 3.53 years ($r = 0.89$), significantly outperforming both the tabular baseline (MAE = 5.12 years, $r = 0.76$) and DINOv3 (MAE = 4.27 years, $r = 0.84$) (Fig. 2a). LeDXA also predicted physiological metrics spanning eight domains (Fig. 2b). These included routine circulating blood biomarkers, such as creatinine (renal; $r = 0.73$), hemoglobin (hematological; $r = 0.68$), high-density lipoprotein (HDL) cholesterol (lipid; $r = 0.57$), glucose (glycemic; $r = 0.35$), and white-blood-cell count (inflammatory; $r = 0.31$). Additionally, the model captured systemic physical and structural measures like the apnea–hypopnea index (AHI, respiratory; $r = 0.54$), heart rate (cardiovascular; $r = 0.43$), and liver fat (hepatic; $r = 0.42$). LeDXA exceeded DINOv3 on all eight biomarkers (all adjusted $P < 0.05$) and outperformed the tabular baseline on five of the eight, tying on HDL cholesterol, glucose, and white-blood-cell count. Together, these results demonstrate that domain-focused pretraining extracts systemic physiological signals more effectively than aggregate DXA measurements or general-purpose encoders.

We then investigated whether the signal captured by LeDXA extends to disease states. We evaluated 37 self-reported chronic conditions in the internal HPP cohort and 28 International Classification of Diseases, 10th Revision (ICD-10) conditions in the external UKBB cohort (n = 25,945) that met a minimum prevalence threshold of 100 cases (Methods), applying the frozen HPP-trained encoder. LeDXA was benchmarked against two baselines: DINOv3 (applied to the same images) and a DXA-tabular baseline utilizing scanner-derived measurements (318 features in HPP, 117 in UKBB; see Methods). All models included age, sex, and BMI as covariates.

Against the DXA-tabular baseline, LeDXA outperformed in 12 of 37 HPP conditions (32%) and 9 of 28 UKBB endpoints (32%), spanning metabolic, musculoskeletal, hematological, respiratory, and sleep-related disorders, and was lower in only one condition per cohort (headache in HPP, hypothyroidism in UKBB), with no significant difference elsewhere. Against DINOv3, it outperformed in 12 of 37 HPP conditions (32%) and 13 of 28 UKBB endpoints (46%), was lower in two HPP conditions (asthma, episodic vertigo) and one UKBB endpoint (celiac disease), and did not differ significantly in the rest (adjusted $P < 0.05$; Tables S2, S3).

Among conditions diagnosed in both cohorts (Fig. 2c), LeDXA improved over the tabular baseline with gains in expected cardiometabolic and musculoskeletal domains, including diabetes (HPP +0.070, UKBB +0.010) and osteoarthritis (HPP +0.026, UKBB +0.018). More notable were systemic conditions not conventionally linked to body-composition readouts, where LeDXA retained cross-cohort value over the tabular baseline for anxiety (HPP +0.013, UKBB +0.026), hyperthyroidism (HPP +0.019, UKBB +0.013), and anemia (HPP +0.010, UKBB +0.017). This advantage extended to cohort-specific conditions (Fig. 2d, e): within HPP, metabolic dysfunction-associated steatotic liver disease (MASLD) (+0.029) and sleep apnea (+0.021); within UKBB, asthma (+0.032) and chronic obstructive pulmonary disease (COPD) (+0.031). Grouping diagnoses into organ-system categories (Methods) confirmed gains over the tabular baseline across multiple domains, driven by cardiovascular, hematological, metabolic, neurologic, rheumatological, and urological categories, consistent with the spatial embeddings carrying prognostic signal beyond the bone and fat readouts DXA conventionally provides (Fig. S4; Full results in Table S4).

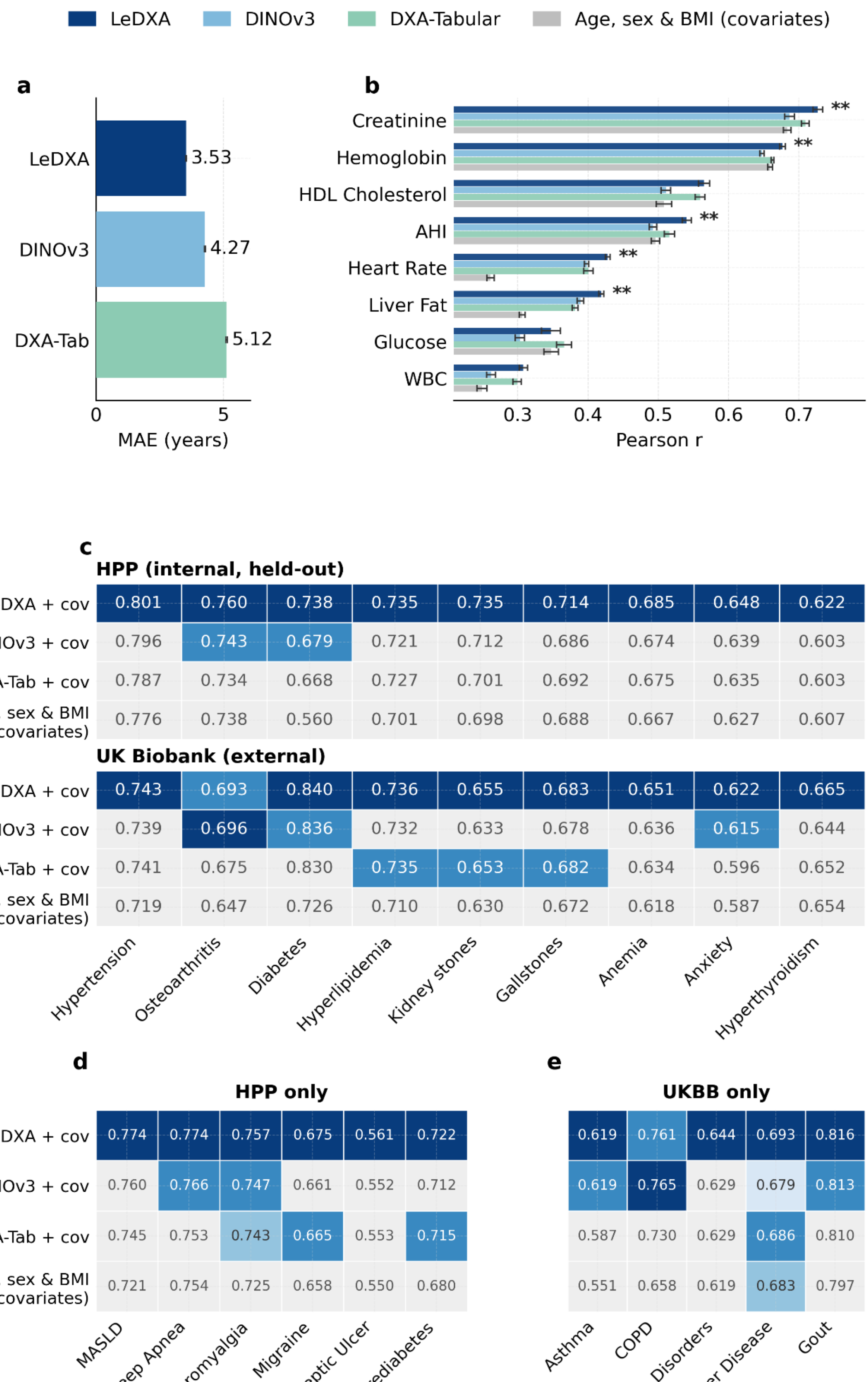

LeDXA
DINOv3
DXA-Tabular
Age, sex & BMI (covariates)
a
LeDXA
DINOv3
DXA-Tab
3.53
4.27
5.12
0
5
MAE (years)
b
Creatinine
Hemoglobin
HDL Cholesterol
AHI
Heart Rate
Liver Fat
Glucose
WBC
**
0.3
0.4
0.5
0.6
0.7
Pearson r
c
HPP (internal, held-out)
LeDXA + cov
DINOv3 + cov
DXA-Tab + cov
Age, sex & BMI (covariates)
0.801 0.760 0.738 0.735 0.735 0.714 0.685 0.648 0.622
0.796 0.743 0.679 0.721 0.712 0.686 0.674 0.639 0.603
0.787 0.734 0.668 0.727 0.701 0.692 0.675 0.635 0.603
0.776 0.738 0.560 0.701 0.698 0.688 0.667 0.627 0.607
UK Biobank (external)
0.743 0.693 0.840 0.736 0.655 0.683 0.651 0.622 0.665
0.739 0.696 0.836 0.732 0.633 0.678 0.636 0.615 0.644
0.741 0.675 0.830 0.735 0.653 0.682 0.634 0.596 0.652
0.719 0.647 0.726 0.710 0.630 0.672 0.618 0.587 0.654
Hypertension
Osteoarthritis
Diabetes
Hyperlipidemia
Kidney stones
Gallstones
Anemia
Anxiety
Hyperthyroidism
d
HPP only
0.774 0.774 0.757 0.675 0.561 0.722
0.760 0.766 0.747 0.661 0.552 0.712
0.745 0.753 0.743 0.665 0.553 0.715
0.721 0.754 0.725 0.658 0.550 0.680
MASLD
Sleep Apnea
Fibromyalgia
Migraine
Peptic Ulcer
Prediabetes
e
UKBB only
0.619 0.761 0.644 0.693 0.816
0.619 0.765 0.629 0.679 0.813
0.587 0.730 0.629 0.686 0.810
0.551 0.658 0.619 0.683 0.797
Asthma
COPD
Sleep Disorders
Liver Disease
Gout

**Figure 2 | LeDXA predicts prevalent disease and physiological traits. a,** Chronological-age prediction in HPP. Models were fitted using ridge regression. Bars show mean absolute error (MAE, years) across 10 random held-out splits, and error bars denote standard error. **b,** Prediction of eight continuous physiological biomarkers in HPP. Models were fitted as in **a**. The covariate-only baseline used age, sex, and BMI alone. Bars show mean Pearson correlation across 10 splits. Asterisks indicate LeDXA significantly outperformed all competing models (two-sided Wilcoxon signed-rank test). **c,** Disease-classification area under the receiver operating characteristic curve (AUROC) for chronic conditions present in both the internal (HPP) and external (UKBB) cohorts, using the frozen HPP-trained encoder. All models were covariate-adjusted for age, sex, and BMI (Methods). Values show mean AUROC across 10 random splits. Within each endpoint, the winning model is shown in dark blue; models not significantly outperformed by the winner remain in lighter blues by rank, while significantly worse models are grayed out (two-sided Wilcoxon signed-rank test, FDR-adjusted $P < 0.05$) **d,** Disease-classification AUROC for HPP-specific conditions. **e,** Disease-classification AUROC for UKBB-specific conditions. In **d** and **e**, models, color scale, split structure, and significance criteria are as in **c**.

As a sensitivity analysis, we also fine-tuned both image encoders end to end on eight representative targets. The relative performance pattern was preserved, with LeDXA remaining consistently above DINOv3 (Fig. S5).

### Longitudinal risk prediction of incident disease

To determine whether LeDXA representations encode prognostic signals beyond cross-sectional phenotypes, we fit penalized Cox proportional-hazards models adjusted for age, sex, and BMI on longitudinal external UK Biobank data (median follow-up 4.3 years; Methods). LeDXA achieved a higher concordance index (C-index) than the tabular DXA baseline for multiple incident disease endpoints (9 of 20 tested endpoints; all reported improvements are FDR-adjusted $P < 0.05$; Fig. 3a and Table S5). The largest predictive gains over the tabular baseline were driven by arthrosis endpoints, specifically hip arthrosis (C-index 0.758 vs. 0.633; $\Delta C = +0.125$) and knee arthrosis (0.744 vs. 0.677; $\Delta C = +0.067$). T2D also demonstrated an improvement (0.753 vs. 0.721; $\Delta C = +0.032$), alongside more modest gains for conditions such as COPD, hypothyroidism, and ischemic heart disease. For incident osteoporosis, LeDXA performed equivalently to the tabular baseline (ns). Alongside the tabular baseline, we concurrently evaluated the general DINOv3 model. While LeDXA and DINOv3 performed equivalently across most incident endpoints, including T2D, LeDXA significantly outperformed DINOv3 specifically in joint diseases, highlighting the distinct domains where DXA-specific pretraining provides an advantage.

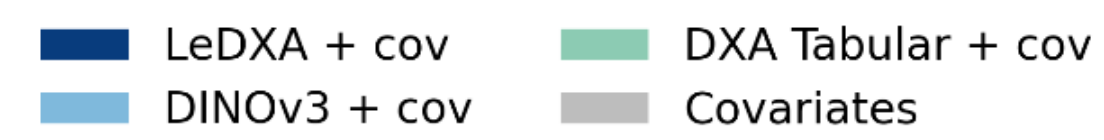

LeDXA + cov
DINOv3 + cov
DXA Tabular + cov
Covariates

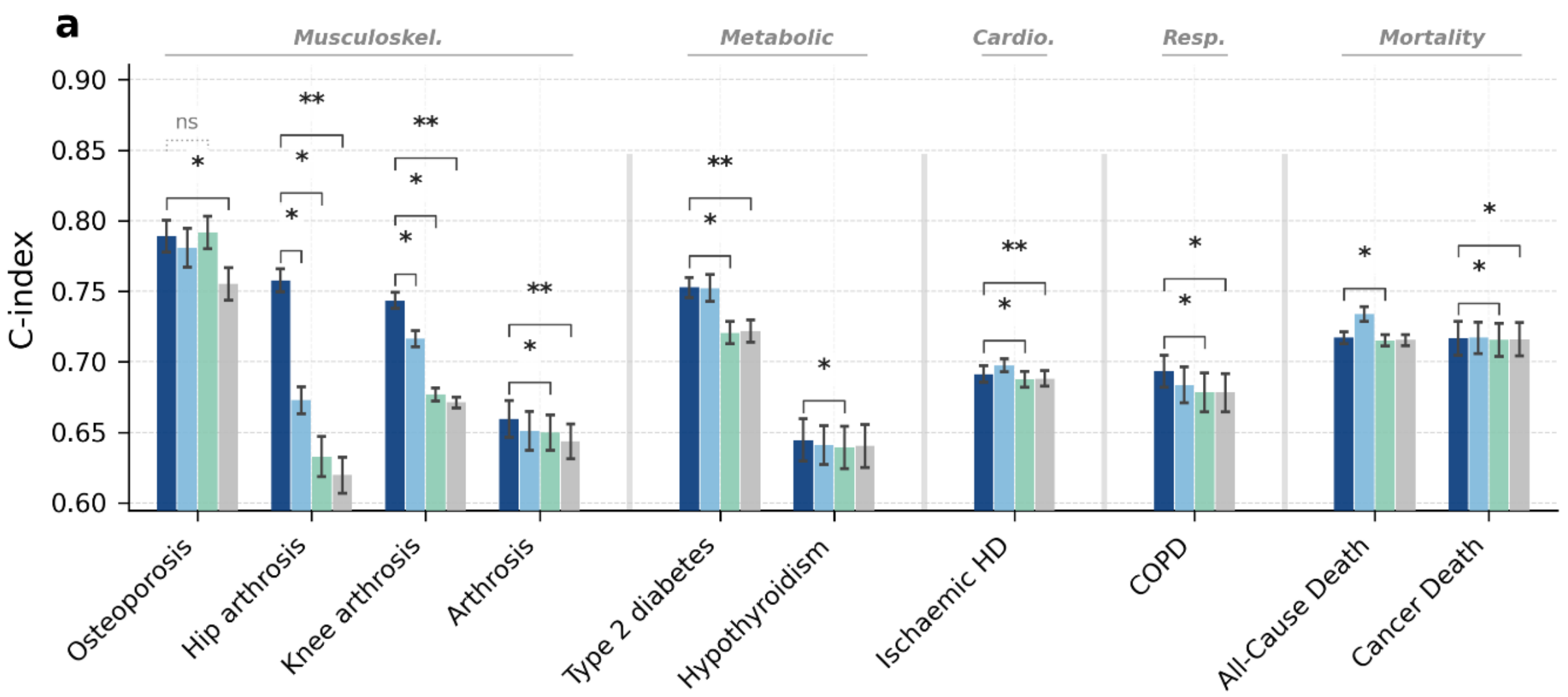

a
Musculoskel.
Metabolic
Cardio.
Resp.
Mortality
C-index
Osteoporosis
Hip arthrosis
Knee arthrosis
Arthrosis
Type 2 diabetes
Hypothyroidism
Ischaemic HD
COPD
All-Cause Death
Cancer Death

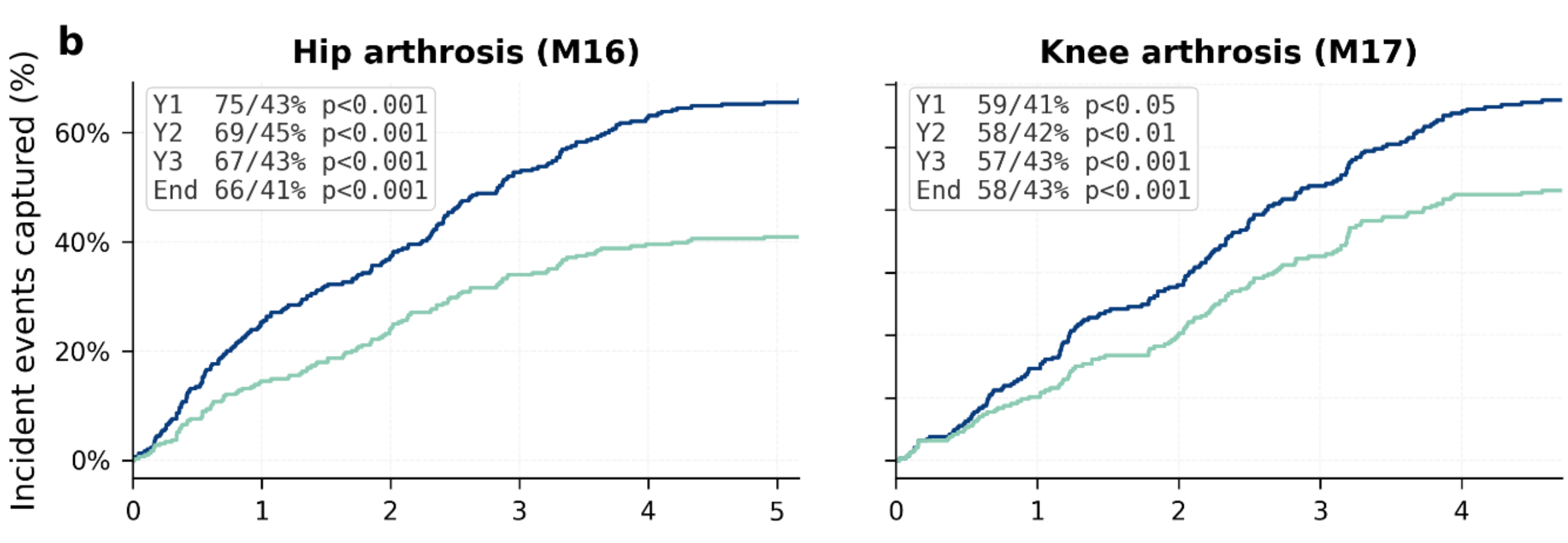

b
Hip arthrosis (M16)
Y1 75/43% p<0.001
Y2 69/45% p<0.001
Y3 67/43% p<0.001
End 66/41% p<0.001
Knee arthrosis (M17)
Y1 59/41% p<0.05
Y2 58/42% p<0.01
Y3 57/43% p<0.001
End 58/43% p<0.001
Incident events captured (%)

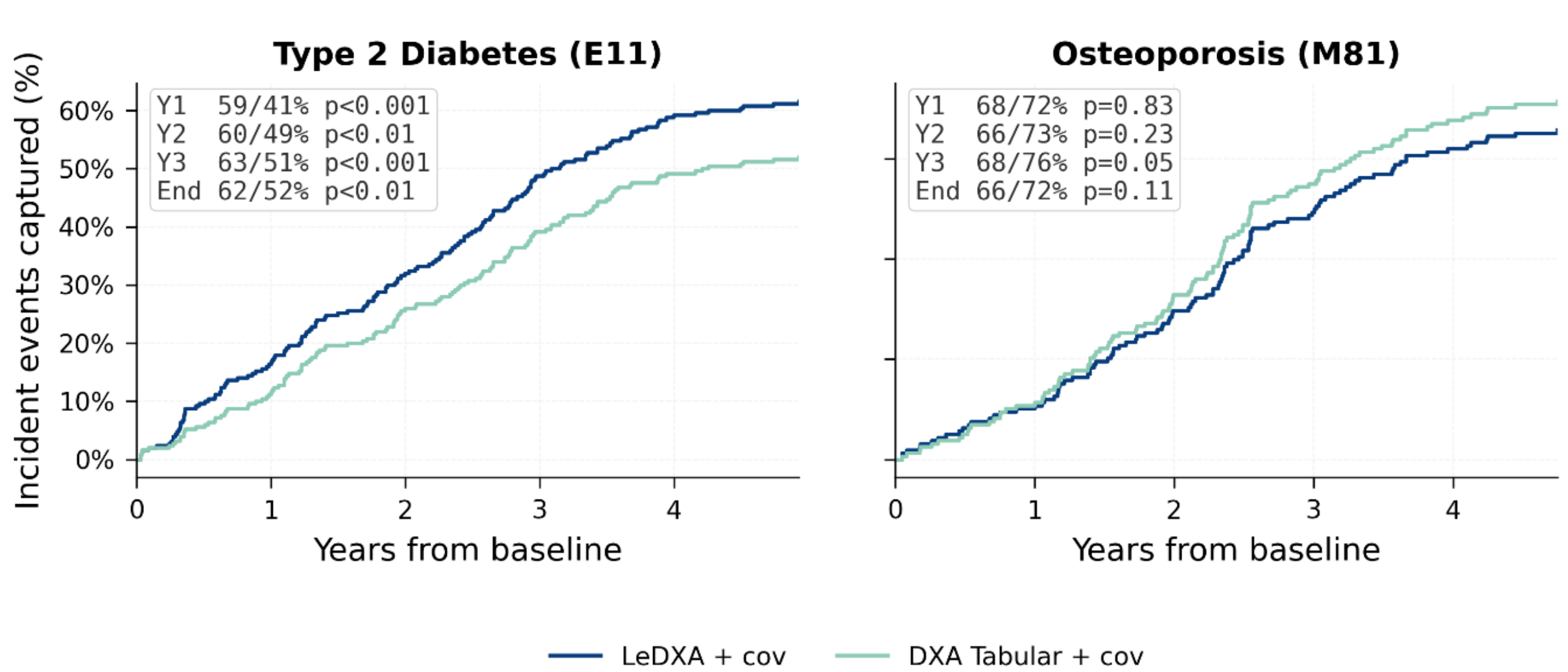

Type 2 Diabetes (E11)
Y1 59/41% p<0.001
Y2 60/49% p<0.01
Y3 63/51% p<0.001
End 62/52% p<0.01
Osteoporosis (M81)
Y1 68/72% p=0.83
Y2 66/73% p=0.23
Y3 68/76% p=0.05
End 66/72% p=0.11
Incident events captured (%)
Years from baseline
LeDXA + cov
DXA Tabular + cov

**Figure 3 | LeDXA improves longitudinal prediction of incident disease in an external cohort.** **a,** Discrimination of incident disease from baseline DXA scans. The plot displays 10 endpoints: the 9 conditions where LeDXA outperformed the DXA-tabular baseline (FDR-adjusted $P < 0.05$) out of 20 tested ICD-10 endpoints, alongside osteoporosis. Cox proportional-hazards models were trained using frozen LeDXA embeddings, DINOv3 embeddings, scanner-derived tabular DXA features, or covariates alone. All models (except the covariate-only baseline) included age, sex, and BMI as covariates. Participants with prevalent disease at baseline were excluded for each endpoint. Bars show the mean concordance index (C-index) across 10 random splits, and error bars denote standard error. Significance brackets indicate pairwise comparisons with LeDXA using two-sided Wilcoxon signed-rank tests with Benjamini–Hochberg FDR correction (*FDR-adjusted $P < 0.05$, **FDR-adjusted $P < 0.01$, ns = not significant). Endpoints are grouped by organ-system category. **b,** Cumulative case capture in the highest predicted-risk quartile (Q4), assigned at baseline, over follow-up, for four representative endpoints: hip arthrosis, knee arthrosis, type 2 diabetes, and osteoporosis. Lines show the cumulative fraction of incident events captured in Q4 for LeDXA (blue) and tabular DXA (green), both including covariates. Insets report the sensitivity of each model at years 1, 2, 3, and at the end of follow-up. p-values compare Q4 sensitivity between models using paired bootstrapping at each timepoint

Because ranking accuracy alone does not establish clinical utility, we asked how many future cases the model's baseline stratification would flag. We stratified participants into risk quartiles at baseline and for each follow-up horizon, measured the fraction of incident cases up to that horizon who had been placed in the top risk quartile (Fig. 3b). We evaluated this for the endpoints with the largest C-index gains, alongside osteoporosis. Hip arthrosis showed the largest and earliest separation: at year 1, LeDXA had flagged 75% of incident cases in its top quartile versus 43% for the tabular baseline, a gap sustained to the end of follow-up (66% vs. 41%; $P < 0.001$). Knee arthrosis showed a similar sustained advantage (59% vs. 41% at year 1, $P < 0.05$; 58% vs. 43% by end of follow-up, $P < 0.001$), and T2D a stable advantage across follow-up (59% vs. 41% at year 1, $P < 0.001$; 62% vs. 52% by end, $P < 0.01$). For osteoporosis, LeDXA matched the tabular baseline at every timepoint, with no significant difference between them (68% vs. 72% at year 1, $P = 0.83$; 66% vs. 72% by end, $P = 0.11$). This is expected, as in the prevalent-disease analysis: the tabular-DXA contains the BMD measurement that defines the diagnosis, so LeDXA recovers this diagnostic signal from the image alone.

## LeDXA-derived representations match known genetic signals in an external cohort

We sought to understand whether the representations learned by LeDXA align with expected genetic signals for body-composition and bone traits. As ground truth, we used gold-standard tabular features from UKBB DXA scans, over which genetic signals have been established [37], and conducted linear-additive GWAS on each phenotype. We then passed raw UKBB DXA scans through the trained model to extract embeddings. We then projected the high-dimensional embeddings onto top 20 principal components (embedding PCs). We performed GWAS on each embedding PC, adjusting for age, sex, and population stratification (Methods), and applied an identical pipeline to DINOv3 embeddings of the same scans. After deduplication by SNP, 1,540 unique SNPs reached genome-wide significance ($P < 5\times10^{-8}$), mapping by position to 1,356 genes (Fig. 4a) over all three sources (LeDXA, DINO, and tabular). LeDXA and the tabular readout shared 187 SNPs, whereas DINOv3 and the tabular readout shared only 36 (Fig. 4b). Genes containing the top hits in LeDXA were established correlates of three biological themes: adiposity, bone and skeletal growth, and neuro-metabolic regulation of body composition. Adiposity and body-fat distribution loci included the canonical FTO (rs1421085, $p < 1e-11$) [38], COBLL1 (rs10195252, $P < 1e-12$) [38] and the CCDC92/DNAH10 [39] locus on 12q24.31 (rs4765127, $P < 1e-10$). Bone- and skeletal-growth loci were led by the CPED1-WNT16 [40] region (rs2707466, $P < 1e-9$) and the adjacent FAM3C [41] signal (rs917727, $P < 1e-16$), with further contributions from RSPO3 (rs1892172, $P < 1e-11$) [38] and CCDC91 (rs11049411, $P < 1e-8$) [39]. Pre-existing signals for neuro-metabolic regulators of body composition were led by APOC1 (rs4420638, $P < 1e-9$) [42] (Table S6). LeDXA also identified 65 single-nucleotide polymorphisms (SNPs) that did not reach genome-wide significance for any tabular trait or DINOv3 embedding. To define independent genetic signals within each source, we performed linkage disequilibrium-based (LD) clumping (PLINK, $r^2 < 0.1$, 500 kb window) separately for tabular, LeDXA, and DINOv3, then compared the resulting loci for genomic overlap across sources. We found 18 loci that reached significance only in the LeDXA embedding space, and not in the GWAS of standard tabular DXA phenotypes or DINOv3 embeddings. Assigning each locus to the primary domain of its lead SNP's (or its high-LD proxy) catalogued associations, most mapped to body-composition, bone, or height traits, with the remainder distributed across other organ systems (Fig. 4c).

To assess the proportion of embedding-space variance explainable by genetics overall, we estimated per-PC SNP-heritability by LD-score regression. Results followed the same rank order (Fig. 4): mean $h^2$ was 0.345 across the tabular traits, 0.143 across the 20 LeDXA components, and 0.098 across the 20 DINOv3 components. The LeDXA embedding PCs exhibited significantly higher SNP heritability than the corresponding DINOv3 embedding PCs (Welch's two-sided t-test, $P = 0.025$). Overall, our results indicate that the domain adaptation provided by LeJEPA leads to much richer genetic alignment in a representation learning setting than SOTA VFMs, recovering known loci and approaching the heritability of hand-curated clinical traits, without explicitly training on genetic data of any kind.

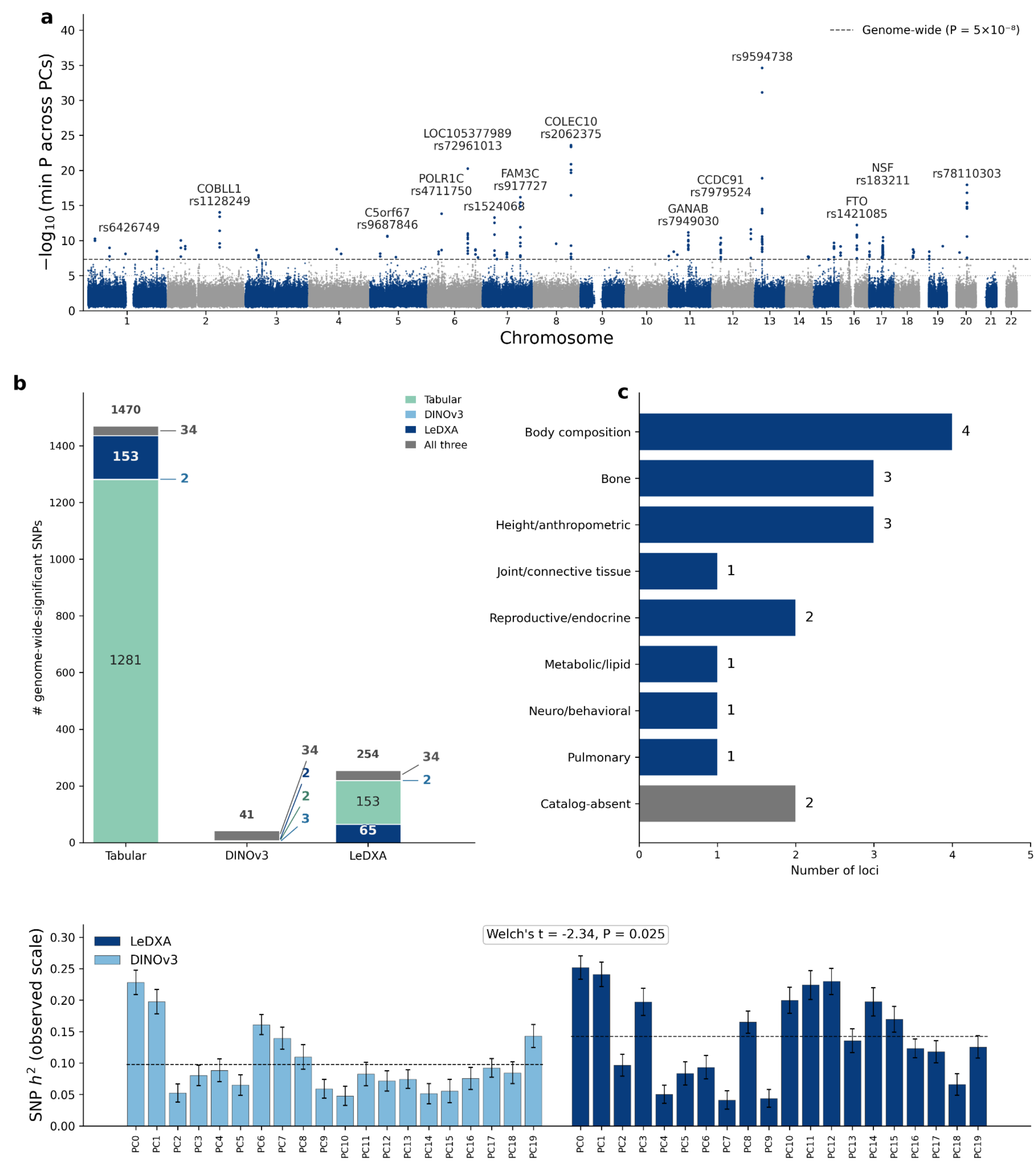

**Figure 4 | LeDXA embeddings capture heritable common-variant architecture and recover established body-composition and skeletal loci.** Genome-wide association studies (GWAS) were conducted in the UK Biobank on the leading 20 principal components (PCs) of the frozen image embeddings (LeDXA and DINOv3) and on scanner-derived tabular DXA traits. **a,** Manhattan plot for the LeDXA embedding GWAS, showing the minimum association P-value per genetic variant across the 20 PCs. Dashed line denotes genome-wide ($P = 5 \times 10^{-8}$) significance threshold. **b,** Number of genome-wide-significant single-nucleotide polymorphisms (SNPs; $P < 5 \times 10^{-8}$) by tabular traits, DINOv3, and LeDXA. Stacked segments partition the identified SNPs into method-specific and shared sets. **c,** Primary GWAS-Catalog trait domain for each of the 18 LeDXA-specific loci (500-kb LD clumping of the 65 LeDXA-specific genome-wide-significant SNPs). Each locus was assigned a single organ-system domain from the catalogued associations of its lead SNP — with 1000 Genomes European LD proxies ($r^2 \geq 0.8$) used only

where the lead SNP was itself uncatalogued — preferring its strongest association; the two loci with no catalogued association in any domain are shown separately as "Catalog-absent." **d,** SNP heritability (LD-score regression, observed scale; $h^2$) for the 20 PCs from LeDXA and DINOv3. Error bars indicate LD-score-regression standard error, and dashed lines represent the mean per method (Welch's two-sided t-test).

## LeDXA-derived biological age predicts clinical outcomes and tracks pharmacological responses

Since LeDXA accurately predicted chronological age in both HPP and UKBB (Fig. 5a,b), we tested whether deviations from the expected age-prediction pattern encoded health-relevant information. We regressed out-of-fold (OOF) predicted age on chronological age and chronological age squared, and used the residual as the biological-age gap [43] (see Methods). Positive gap values indicate DXA morphology that appears older than expected for a participant's chronological age and a negative gap indicates a younger DXA morphology. Empirically, the gap showed no correlation with age overall ($r = -0.002$) and within each sex (females $r = +0.006$, males $r = -0.019$; Fig. S6).

The DXA biological-age gap captured a sex-divergent aging process. Comparing the highest (Q4, biologically older) and lowest (Q1, biologically younger) age-gap quartiles within UKBB, we found distinct profiles for each sex (Fig. 5c, d). In females, higher biological age was dominated by lower BMD and bone mineral content (BMC), with modest increases in VAT. In males, it was marked by generalized ALM loss and widespread BMD reductions (arms, legs, and femoral sites), alongside a paradoxical rise in lumbar BMC and BMD. Notably, this lumbar rise most likely reflects a known DXA artifact rather than real bone gain: in older men, degenerative changes, aortic calcification, and vertebral fractures can falsely raise measured L1–L4 density [44]. The hip and femoral sites, which this artifact affects less, showed the expected decline with age. Because the LeDXA age gap reproduces these known, site-specific patterns of aging, it confirms that the embeddings extract true anatomical structure from the image rather than relying on non-biological artifacts (this pattern is also replicated in the internal HPP cohort; Fig. S7).

These structural changes were accompanied by clinically relevant differences. Consistent with the observed ALM deficit, low muscle quantity, defined using the European Working Group on Sarcopenia in Older People 2 (EWGSOP2) [45] appendicular lean mass index (ALMI) thresholds of $<7.0$ kg/m² in men and $<5.5$ kg/m² in women, rose monotonically across UKBB biological-age-gap quartiles. Prevalence increased from 6.6% in Q1 to 13.6% in Q4 among men (PR 2.06, 95% CI 1.62–2.63, $P = 1.5\times10^{-9}$) and from 7.0% to 10.2% among women (PR 1.45, 95% CI 1.14–1.83, $P = 1.9\times10^{-3}$). Each additional year of biological-age gap was associated with a 10.5% increase in the odds of low ALMI in men and a 6.3% increase in women, independent of chronological age.

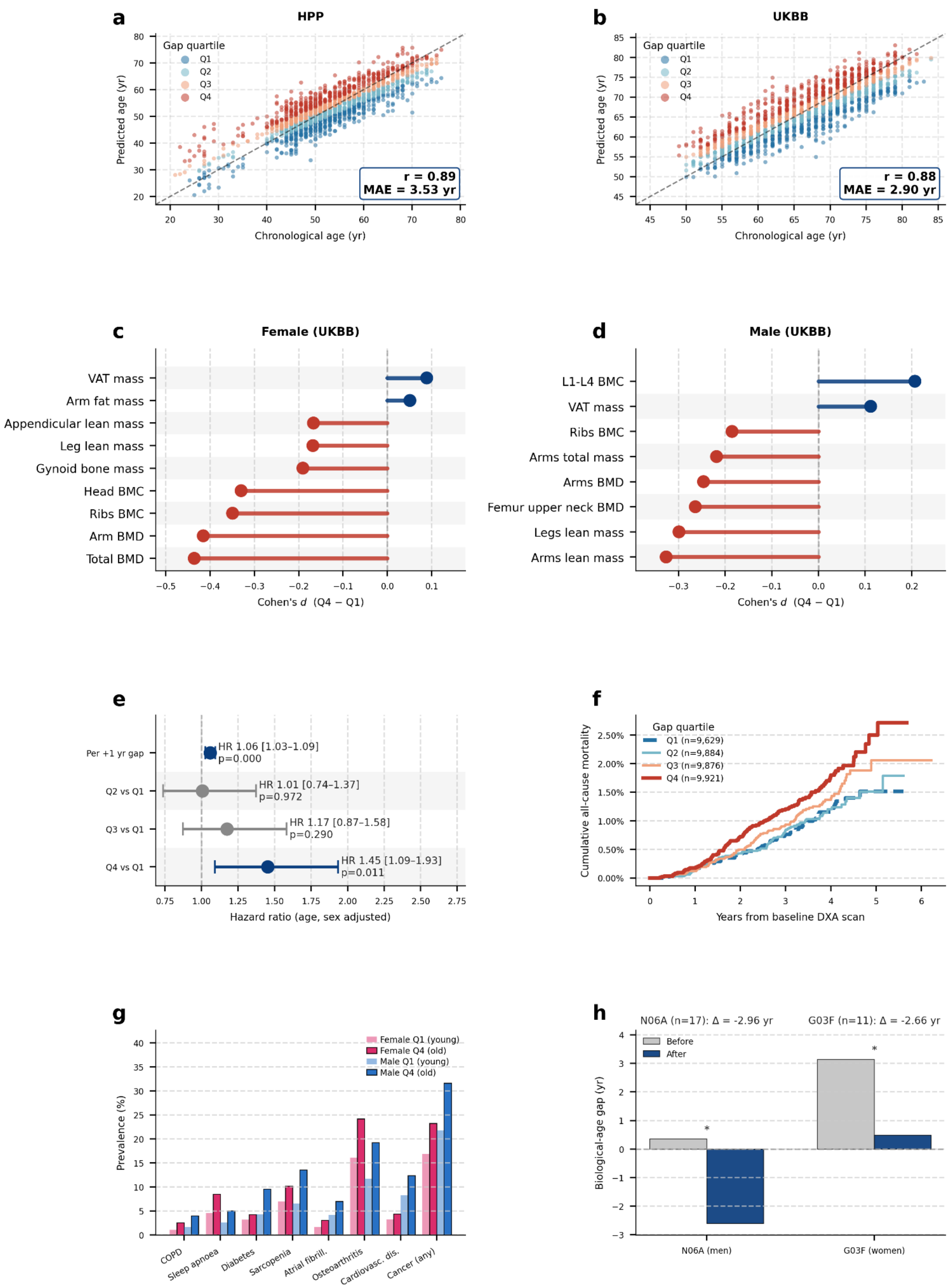

**Figure 5 | LeDXA-derived biological-age gap captures sex-specific morphology, disease burden, mortality risk, and medication-associated change. a,b,** Chronological-age prediction from LeDXA embeddings in the internal (HPP) and external (UKBB) cohorts. Predicted age was generated out-of-fold. The biological-age gap was defined as the residual after regressing predicted age on

chronological age and chronological age squared. Points are colored by biological-age-gap quartile, with Q1 representing younger-appearing and Q4 representing older-appearing DXA morphology. Dashed lines indicate the identity line; boxed annotations report the Pearson correlation coefficient ($r$) and mean absolute error (MAE). **c,d,** Sex-stratified DXA phenotypes associated with biological-age-gap quartiles in UKBB. Lollipop plots show Cohen's d comparing Q4 with Q1 (positive values indicate higher measurements in Q4). Displayed phenotypes represent the DXA-derived traits with the largest absolute effect sizes per sex (all displayed associations FDR-adjusted $P < 0.05$). **e,** Age- and sex-adjusted Cox proportional-hazards models for all-cause mortality in the UKBB mortality analytic cohort (n = 39,310; 377 deaths; median follow-up, 3.40 years). Points indicate hazard ratios (HRs) for each quartile relative to Q1, and horizontal lines indicate 95% confidence intervals (CIs). **f,** Cumulative all-cause mortality in the same UKBB cohort stratified by baseline biological-age-gap quartile. **g,** Prevalence of selected diseases in UKBB by sex and biological-age-gap quartile. Bars display the eight conditions with the largest mean prevalence ratio (PR) between Q4 and Q1 among the 17 evaluated conditions that reached significance (FDR-adjusted $P < 0.05$). **h,** Paired before/after analysis of biological-age-gap changes following the first reported initiation of significant Anatomical Therapeutic Chemical (ATC) level-3 medication classes in HPP participants with serial DXA scans.

Beyond measures of musculoskeletal quality, this accelerated morphological aging translated directly into a broadly elevated systemic disease burden. Across a panel of 17 prevalent ICD-10/self-reported conditions evaluated at baseline (Methods), 16 of 17 in women and 17 of 17 in men were significantly more prevalent in Q4 than Q1 (adjusted $P < 0.05$; Fig. 5g and Fig. S8), spanning musculoskeletal, cardiometabolic, respiratory, and oncologic domains. The strongest gradients were observed in COPD (women PR 2.32, men PR 2.43), rheumatoid arthritis (women PR 2.26, men PR 2.28), and T2D. Disaggregating disease categories revealed that the cardiovascular signal is concentrated in heart failure (women PR 1.72, adjusted $P = 0.038$; men PR 2.01, adjusted $P = 2.9\times10^{-5}$) and myocardial infarction (women PR 1.56, adjusted $P = 0.028$; men PR 1.69, adjusted $P = 2.2\times10^{-8}$), and that the OA signal is most pronounced in hip arthrosis (women PR 1.89, adjusted $P = 3.7\times10^{-3}$; men PR 2.69, adjusted $P = 2.6\times10^{-4}$), patterns consistent with the bone- and lean-mass deficits that define the Q4 phenotype.

To test whether the biological-age gap captures prognostic information beyond chronological age, we modelled all-cause mortality across the UKBB data (n = 39,310; 377 deaths over a median follow-up of 3.40 years) using Cox proportional-hazards regression adjusted for age and sex (Methods; Fig. 5e, f). Each additional year of biological-age gap was associated with a 5.9% increase in mortality hazard (HR 1.06, 95% CI 1.026–1.094, $P = 4.5\times10^{-4}$); when modelled categorically, individuals in the oldest gap quartile (Q4) experienced a 45.3% higher hazard than those in the youngest (Q1) (HR 1.45, 95% CI 1.09–1.94, $P = 0.011$), with intermediate quartiles ordered monotonically. Because the model is adjusted for age, this gradient cannot be explained by older subjects accumulating in Q4; instead, it reflects a morphological signal of accelerated physical aging that retains predictive value on top of standard demographic risk factors.

Finally, we asked whether the biological-age gap tracks clinical interventions over time. In HPP, DXA scans are paired with longitudinal self-reported medication questionnaires. Restricting to participants whose first reported use of a drug fell between their baseline and follow-up scans, we tested paired pre/post changes in the gap across 23 Anatomical Therapeutic Chemical level-3 (ATC-3) drug classes (Methods; Fig. 5h and Table S7). After correction, two classes showed a significant gap reduction after initiation: hormone replacement therapy (HRT) in women (G03F; n = 11, Δ = −2.66 yr, adjusted $P = 0.034$) and antidepressants in men (N06A; n = 17, Δ = −2.96 yr, adjusted $P = 0.037$). For antidepressants, BMI rose after initiation (+1.3 kg/m²), consistent with the weight gain reported for some agents [46]; yet the gap reduction tracked lean mass (age-adjusted partial $r = -0.57$, $P = 0.044$), not fat mass ($r = -0.38$, $P = 0.20$) or VAT ($r = -0.001$, $P = 0.998$). For HRT, the gap reduction was directionally associated with preserved lean mass ($r = -0.52$) and total BMD ($r = -0.48$), but not fat mass ($r = 0.03$, $P = 0.95$); although these did not reach significance at the available sample size, the BMD pattern matches the known skeletal effects of menopausal hormone therapy [47].

**Unsupervised LeDXA embeddings identify a lean- and high-bone-density female phenotype associated with favorable physiological profiles**

Uniform manifold approximation and projection (UMAP) [48] visualization of LeDXA embeddings across HPP revealed sex-specific structure: women separated into two visually distinct groups, men did not (Fig. 6a). Because the embeddings were learned without sex labels or clinical endpoints, we asked whether this structure reflected interpretable physiological variation. Clustering the female embeddings (one scan per participant) identified two groups Cluster A (n = 3,162) and Cluster B (n = 1,232).

The clusters differed primarily in overall size (ΔBMI = 4.3 kg/m², d = 1.1) with negligible age separation (Δage = 1.4 years, $d = 0.13$). But many women overlapped in BMI (Fig. 6b), an unadjusted comparison would mostly recover the known effects of higher BMI. We instead asked why women of similar age and size fall into divergent morphological clusters. To isolate body composition from gross size, we matched the clusters 1:1 on age and BMI, yielding 1,035 pairs with balanced covariates (SMD ≤ 0.03; Methods).

At matched age and BMI, 81 of 117 body-composition features differed (adjusted $P < 0.05$; Table S8). Cluster B showed lower peripheral fat (legs $d = -0.94$; gynoid $d = -0.75$) and higher regional lean mass (arms $d = +0.56$; trunk $d = +0.52$), together with higher BMC and BMD (pelvis BMC $d = +0.34$; femoral BMD $d = +0.25$; 111 of 182 bone features differed). Independent of age and BMI, Cluster B therefore defines a lean-mass and high-bone-density axis with reduced gynoid fat.

This profile tracked with systemic physiological advantages in the higher-lean-mass cluster (Fig. 6c). Muscle and metabolic markers were more favorable, including higher grip strength ($d = +0.46$) and higher creatinine (compatible with greater muscle mass; $d = +0.30$). Autonomic markers pointed to higher parasympathetic tone (lower resting heart rate, higher heart-rate variability, lower REM-sleep heart rate; all $|d| \geq 0.31$), and hematological markers followed the same favorable direction (Fig. 6c). The higher-lean-mass cluster also reported healthier behaviors: more frequent moderate activity ($d = +0.28$), higher relative energy intake ($d = +0.20$), and a higher Mediterranean diet score ($d = +0.19$). At matched age and BMI, these differences span muscle, cardiometabolic, autonomic, and behavioral measures, indicating that the embedding separates women along an axis of physiological health that body size alone does not capture.

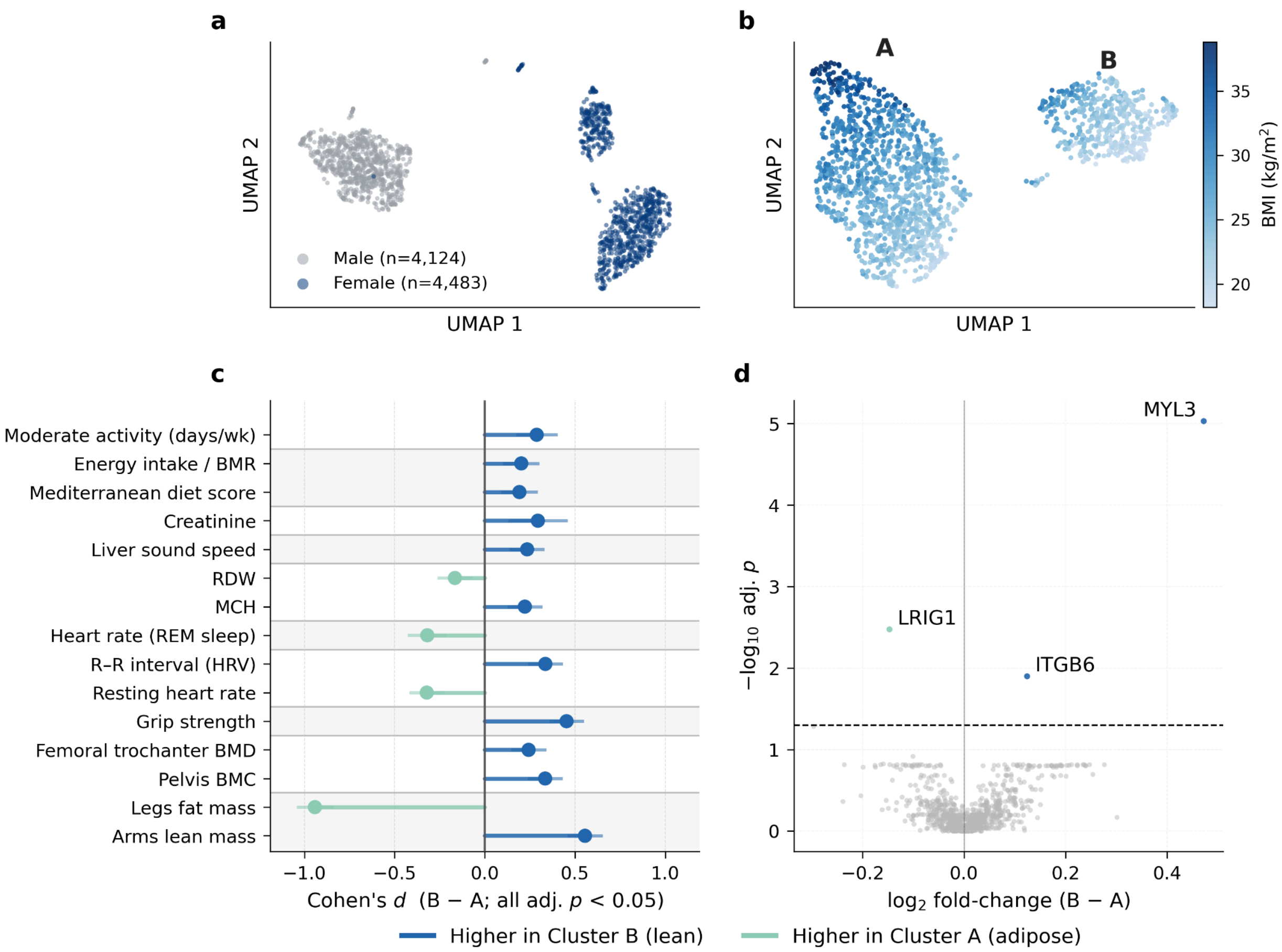

**Figure 6 | Unsupervised clustering of LeDXA embeddings identifies a female body-composition axis. a,** Uniform manifold approximation and projection (UMAP) of whole-body DXA embeddings from the Human Phenotype Project (HPP), colored by sex. **b,** UMAP of female HPP participants colored by body mass index (BMI). **c,** Phenotypic differences between Cluster B and Cluster A following 1:1 Mahalanobis-distance matching on age and BMI (yielding 1,035 matched pairs). Lollipop plots display Cohen's d (Cluster B minus Cluster A), with positive values indicating higher measurements in Cluster B. Displayed associations are significant at FDR-adjusted $P < 0.05$. **d,** Plasma-proteomic differential-abundance analysis in the matched cohort utilizing the Olink platform. The x-axis represents the $\log_2$ fold-change between clusters (Cluster B minus Cluster A), and the y-axis indicates the $-\log_{10}$ (FDR-adjusted P-value).

We next asked whether this phenotype was visible at the molecular level. Across matched-cohort proteomics, metabolomics, microbiome, and transcriptomics, the strongest signal was in plasma proteomics (three Olink proteins differed, adjusted $P < 0.05$; Fig. 6d and Table S9). MYL3, a slow-skeletal and ventricular myosin essential light chain [49], and ITGB6, were higher in the higher-lean-mass cluster ($\log_2$FC = +0.47 and +0.12); LRIG1 was higher in the lower-lean-mass cluster ($\log_2$FC = −0.15). Nuclear magnetic resonance (NMR) metabolomics returned higher creatinine and lower GlycA in the higher-lean-mass cluster, with no differences in transcriptomics or microbiome. Because these platforms are independent of the imaging features that defined the clusters, MYL3 and creatinine, the latter elevated in both NMR and the blood panel, support a true difference in muscle physiology rather than fat distribution alone.

## Discussion

DXA is acquired at population scale primarily for osteoporosis screening and body-composition assessment. However, our findings suggest that raw DXA images contain complex physiological information that standard scalar readouts may not fully capture. By training LeDXA, a compact, self-supervised vision model directly on unannotated scans, we recovered standard densitometric targets while identifying novel clinical and prognostic associations. Standard tabular readouts reduce a patient's scan to regional sums of fat, lean mass, and bone, stripping away the spatial geometry of tissue distribution. By preserving this structure, LeDXA captured health states that traditional measures blur. The value of capturing this spatial structure was initially evident in our cross-sectional analyses, where LeDXA outperformed both the tabular DXA baseline and the generalist DINOv3 model across a wide spectrum of prevalent-disease endpoints and biomarkers. While these cross-sectional results highlighted LeDXA's ability to capture concurrent health states, its most striking advantage emerged in our longitudinal prediction of incident disease. LeDXA's predictive gains over the tabular baseline were largest for hip and knee arthrosis, the endpoints with the biggest effect sizes and the domains where DXA-specific pretraining significantly outperformed DINOv3. This is expected, as OA is a fundamentally structural and biomechanical disease, where spatial alterations in bone shape often precede apparent cartilage loss [50,51]. While recent studies have used supervised, OA-specific classifiers on skeletal imaging to predict joint replacement [52], our findings show that LeDXA extracts these morphological risk factors natively, without any radiographic labels. LeDXA was also a stronger prognostic predictor of incident T2D than the tabular baseline. Increasing evidence demonstrates that the specific spatial distribution of adipose tissue, particularly VAT, is a primary driver of incident T2D [53,54]. Although the tabular baseline already includes scanner-derived VAT, which LeDXA recovered from the raw image with high fidelity ($r = 0.96$), the embeddings still provided significant added predictive value. Standard readouts summarize fat distribution through regional and android-gynoid measures, reducing the scan to aggregate values. The added predictive value of LeDXA suggests that the embeddings capture finer spatial distributional detail relevant to metabolic risk.

These analyses link LeDXA to health outcomes. To test whether its features reflect underlying biology rather than the outcome labels themselves, we turned to genetics, asking whether they are tied to inherited variation and capture more of this signal than a general-purpose model. We ran a GWAS on the embedding principal components in the UKBB, an external cohort the model was never trained on, and compared the results to the genetics of the standard tabular DXA measures. The components mapped to known body-composition and bone loci, showing that part of what LeDXA learns reflects the same genetics that shape the standard measures. This genetic signal was stronger for LeDXA than for DINOv3 (Welch's t, $P = 0.025$), indicating that DXA-specific training captures body-composition biology better than a general model. We hypothesize that this better alignment with heritability of LeDXA may be linked to its added predictive power.

LeDXA also captures continuous measures of systemic risk, such as biological age. The LeDXA biological-age gap stratified all-cause mortality independent of chronological age and sex, indicating that DXA-derived morphology captures axes of structural aging not entirely explained by age alone. We also observed sex-specific structural patterns, with the age gap associated predominantly with lower ALM in men and with lower BMD in women. Furthermore, the model appears sensitive to dynamic physiological shifts. The biological-age gap decreased following self-reported initiation of antidepressants and HRT, which are associated with weight gain and preservation of BMD, respectively [46,47]. The change tracked lean mass rather than fat, the same compartment whose preservation is a central concern in pharmacological weight loss. Together, these findings indicate that LeDXA could potentially serve as a tool for tracking individualized health trajectories.

Clustering of LeDXA embeddings identified a favorable female phenotype characterized by greater lean mass, stronger muscle function, higher bone density, and lower peripheral adiposity at the same age and BMI. This pattern refines the literature on the protective associations of gynoid body composition [55,56]: favorable metabolic and autonomic features were present despite lower gynoid fat, suggesting that lower-body adiposity alone is insufficient to explain this phenotype. Instead, the findings implicate the broader balance between adipose and musculoskeletal compartments. This interpretation was supported by independent molecular measurements: the lean-mass cluster showed higher MYL3, a myosin essential light chain [49], higher creatinine, and lower GlycA, an NMR marker of systemic inflammation [57]; the lower-lean-mass cluster showed higher LRIG1, which promotes adipogenesis via BMP signaling [58].

This fine-grained monitoring of muscle and bone is particularly vital in the era of pharmacological weight loss. While therapies like (Glucagon-Like Peptide-1) GLP-1 receptor agonists can drive profound reductions in body weight, they are also associated with the loss of lean mass [59,60]. Our results align with the perspective that weight-loss interventions should be evaluated not just by total mass reduction, but by the preservation of lean muscle and bone. LeDXA offers a potential framework to monitor these critical tissue compartments, consistent with the development of newer anti-obesity pathways (such as GDF15–GFRAL) that are being pursued to reduce fat with less muscle loss [61].

This study has several strengths. Our analysis drew on a large multimodal cohort of 8,820 participants with DXA imaging linked to deep phenotyping, including multi-omics data, which let us relate learned image features to molecular and clinical measures in the same individuals. By adjusting all baselines for age, sex, and BMI and comparing LeDXA to tabular features from the same scans, we isolated the added value of raw spatial structure. Validation in the UK Biobank further showed that the model transfers to an independent cohort.

However, several limitations warrant consideration. All analyses were observational, and the medication-initiation analyses relied on self-reported data and require replication. Both cohorts are also subject to healthy-volunteer selection: the HPP cohort is volunteer-based and largely healthy, while the UK Biobank, although independent, shares this bias. Performance was not evaluated separately across ancestry groups. The reported effect sizes and risk estimates may therefore not generalize to unselected clinical populations, diverse ancestry groups, or routine clinical DXA settings. Finally, longitudinal follow-up was limited to a median of 4.3 years; longer follow-up is needed to determine how LeDXA captures intra-individual changes in body composition over time.

In summary, LeDXA shows that predictive and heritable representations of body composition can be learned directly from unannotated DXA scans. By utilizing the full spatial structure of the images, LeDXA adds prognostic information beyond standard DXA readouts and provides a scalable framework for advancing our understanding of systemic disease risk, musculoskeletal health, and aging.

# Methods

## Study populations

**Internal cohort.** The Human Phenotype Project (HPP) [35] is a longitudinal deep-phenotyping cohort of generally healthy Israeli adults (individuals with severe medical conditions were excluded at enrolment), with follow-up assessments every two years, designed to capture dense, repeated phenotyping across multiple physiological systems — including anthropometry, blood biochemistry, cardiometabolic and sleep measures, abdominal and vascular ultrasound, and molecular profiling (genomics, transcriptomics, proteomics, metabolomics and gut microbiome). The baseline cohort comprised 13,456 participants. Participants underwent whole-body DXA at baseline on a GE Healthcare Lunar Prodigy Advance densitometer (see Data Acquisition and Preprocessing), with a subset returning at approximately 2, 4 and ≥6 years, enabling assessment of both cross-sectional associations and longitudinal trajectories. 11,540 DXA scans from 8,820 unique participants were used for self-supervised pretraining and internal evaluation; the analytic sample with complete covariates comprised 8,759 participants (mean age 50.3 ± 9.7 years, range 18–80; 52% women; Table S1). The HPP DXA imaging dataset includes all scans acquired up to February 2026

**External cohort.** External validation used the UK Biobank (UKBB) [36], a population-based prospective cohort of approximately 500,000 UK adults (502,244 participants in the data release analyzed here) aged 40–69 years at recruitment (2006–2010), providing diverse representation of age, health status and disease burden. A subset of participants returned for a multi-organ imaging assessment (first imaging visit, Instance 2) that includes whole-body DXA on GE Healthcare Lunar iDXA systems (see Data Acquisition and Preprocessing). We restricted analyses to this first imaging visit and selected whole-body scans in the compatible MONOCHROME2 photometric format that passed image-quality control, yielding 47,400 participants; the analytic sample with complete covariates comprised 45,789 participants (mean age 66.1 ± 7.6 years, range 44–85; 52% women; Table S1). No UKBB data were used at any stage of pretraining; all UKBB analyses were performed on embeddings produced by the frozen HPP-trained model.

**Exclusions and analytic samples.** Beyond each cohort's own enrolment criteria, no additional diagnosis- or medication-based exclusion criteria were applied, so participants were not further filtered by disease status or medication use. Beyond selection on scan format and image quality (degraded or partial scans were excluded; see Data Acquisition and Preprocessing), downstream analyses were restricted to participants with complete age, sex and BMI and, for tabular models, ≤50% missing DXA features (see Feature preprocessing), yielding analytic samples of 8,759 (HPP) and 45,789 (UKBB), whose baseline characteristics are summarized in Table S1. For the longitudinal incident-disease models, participants with the relevant condition present at or before imaging baseline were additionally excluded on a per-endpoint basis (see Longitudinal Incident-Disease Risk). Participant flow from cohort selection to the analytic sample is shown for both cohorts in Fig. S2.

**Ethics.** The Human Phenotype Project was approved by the Institutional Review Board (IRB) of the Weizmann Institute of Science (reference number 1719-1), and all HPP participants provided written informed consent. UK Biobank has ethical approval from the North West Multi-center Research Ethics Committee (as a Research Tissue Bank), and all UK Biobank participants provided written informed consent; the present analyses were conducted under UK Biobank Application Number 28784.

**Data acquisition and preprocessing**

Whole-body DXA acquires attenuation at two X-ray energies, allowing separation of bone mineral from soft tissue and partition of soft tissue into lean and fat compartments. Raw DXA images were exported in DICOM format. In the internal cohort (HPP), scans were acquired on a GE Healthcare Lunar Prodigy Advance densitometer; in the external cohort (UKBB), scans were acquired on GE Healthcare Lunar iDXA systems. In both cohorts, two image channels were extracted per scan: a bone-modality image (reflecting bone mineral distribution) and a tissue-modality image (reflecting the spatial distribution of adipose and lean soft tissue).

Native image dimensions varied substantially: HPP images ranged from 128–413 pixels in height by 44–124 pixels in width (with frequent degraded or partial scans at the lower extremes), whereas UKBB images ranged from 280–1,192 pixels in height by 216–432 pixels in width. The most common dimensions were 412×124 for HPP (12.8% of scans) and 811×272 for UKBB (12.1% of scans). All images were resized to a uniform 384×128 pixels using bicubic interpolation. Pixel intensities were normalized to zero mean and unit variance using cohort-specific training-set statistics (HPP: $\mu = 0.196$, $\sigma = 0.284$; UKBB: $\mu = 0.189$, $\sigma = 0.280$). Each single-channel image was replicated across three channels to match the encoder input. Regional bone scans of the left femur, right femur, and lumbar spine (L1–L4) were extracted using scanner-reported region-of-interest (ROI) bounding boxes when available.

**Self-supervised pretraining with LeJEPA**

**Architecture.** LeDXA is a Vision Transformer [62] (ViT-Small/16; embedding dimension 384, depth 12, 6 attention heads; 21.7 M parameters) followed by a three-layer MLP projection head (384 → 2,048 → 2,048 → 64, with batch normalization and ReLU between layers; Fig. S1). The model was trained from random initialization entirely on HPP DXA scans, with no ImageNet pretraining or external weights, using the LeJEPA framework [34].

**View construction.** Each training sample had 10 augmented views per modality stream: 2 global views from the full-body image and 8 local views. Global views were produced by random resized crop to 384×128 pixels (scale 0.5–1.0, aspect ratio 0.25–0.45, bicubic interpolation), random rotation (±10°), random horizontal flip (applied with probability 0.5), Gaussian blur (kernel 5, $\sigma =$ 0.1, probability 0.5), and color jitter (brightness and contrast 0.4, probability 0.5). Local views were generated at 96×96 pixels. Let $K \in \{0, 1, 2, 3\}$ denote the number of available anatomical ROI scans (left femur, right femur, lumbar spine L1–L4). The K available ROI scans were augmented with random resized crop (scale 0.6–1.0, aspect ratio 0.7–1.3), random rotation (±15°), horizontal flip (probability 0.5), and color jitter (probability 0.5); the remaining $8 - K$ local views were synthesized as small random crops of the full-body image (scale 0.1–0.3, aspect ratio 0.7–1.3, bicubic), with horizontal flip (probability 0.5) and color jitter (probability 0.5). The bone and tissue modalities were each passed through this view-construction pipeline, exposing the model jointly to whole-body composition and fine-grained regional skeletal structure within a shared embedding space.

**LeJEPA objective.** Let the encoder $f$ and projection head $g$ map a view x to a projection $z = g(f(x)) \in \mathbb{R}^d$ ($d = 64$). For a sample with $G = 2$ global views and $L = 8$ local views ($V = G + L$), every view is trained to predict the centroid $\mu$ of the global-view projections, with the squared deviation averaged over views and the d embedding coordinates:

$$\mathcal{L}_{\text{pred}} = \frac{1}{V}\sum_{v=1}^{V}\frac{1}{d}\|z_v - \mu\|_2^2, \qquad \mu = \frac{1}{G}\sum_{v=1}^{G} z_v.$$

**SIGReg regularization.** To prevent representational collapse, LeJEPA drives the projection distribution toward an isotropic standard Gaussian with Sketched Isotropic Gaussian Regularization (SIGReg). SIGReg draws $S = 2{,}048$ random unit directions (resampled every step), projects the batch onto each to form one-dimensional slices, and averages the Epps–Pulley goodness-of-fit statistic over the slices. For a slice with empirical characteristic function $\hat{\varphi}(t)$, that statistic is the squared distance to the standard-normal characteristic function $\psi(t) = \exp(-t^2/2)$, weighted by $\psi$:

$$\mathcal{L}_{\text{SIGReg}} = \frac{1}{S}\sum_{s=1}^{S} B \int |\hat{\varphi}_s(t) - \psi(t)|^2\, \psi(t)\, \mathrm{d}t,$$

where B is the batch size and the integral is evaluated by a 17-knot trapezoidal rule over

$t \in [0, T]$, $T = 5$ (halved by the integrand's symmetry).

The total self-supervised objective combines the two terms, $\mathcal{L} = (1-\lambda)\,\mathcal{L}_{\text{pred}} + \lambda\,\mathcal{L}_{\text{SIGReg}}$, with $\lambda = 0.05$; SIGReg is evaluated per view and averaged, and the loss is computed independently for the bone and tissue view streams and averaged.

**Training configuration**

LeDXA was optimized with AdamW [63] (learning rate $5\times10^{-4}$, weight decay $5\times10^{-2}$) under a cosine-annealing schedule decaying to zero with a 30-epoch linear warmup, for 400 epochs on a single NVIDIA L40S GPU with batch size 256, completing in approximately 10 hours. Weight decay was applied to all parameters. Stochastic-depth regularisation (drop-path rate 0.1), bfloat16 mixed-precision training, and gradient clipping (max norm 1.0) were used throughout. LeDXA was implemented in PyTorch 2.6 (CUDA 12.4) with timm 1.0.26 and torchvision 0.21.

**Embedding extraction.** After pretraining the encoder weights were frozen. For every scan, bone- and tissue-modality embeddings were extracted as the class token [CLS] from the final transformer layer of the encoder, using the deterministically resized 384×128 image (no augmentation); a fused representation was formed by concatenating the two modality embeddings. The same frozen HPP-trained model was applied without modification to UKBB scans for all external-validation analyses.

**Baseline models**

**DINOv3.** A DINOv3 ViT-H+ encoder (≈860 M parameters) pretrained on curated natural images (LVD-1698M and ImageNet) under the DINOv3 framework [23] was used as a general-purpose vision baseline. DXA images were resized and normalized using the model's native preprocessing, and frozen final-layer [CLS] features were extracted under the same bone/tissue/fusion protocol as LeDXA.

**DXA tabular features.** Machine-derived measurements automatically generated by the DXA scanner software were modeled under the same statistical protocols as the frozen embeddings. In HPP, 318 tabular DXA features were available (Table S10), encompassing whole-body and regional BMD, BMC, regional fat, lean mass, VAT mass and structural parameters (area, width, height). In the UKBB cohort, 117 features were available after filtering to exclude administrative metadata (scanner details, measurement dates, assessment center identifiers) (Table S10), reflecting differences in scanner software versions and data-release conventions.

**Feature preprocessing.** For the downstream tabular- and embedding-based models (regression, classification and Cox analyses), subjects with >50% missing values in the tabular feature set were excluded; remaining missing values were median-imputed and all features z-score standardized, $z_i = (x_i - \mu_{\text{train}})/\sigma_{\text{train}}$, using training-fold statistics only (applied unchanged to the validation fold) to prevent leakage. This pipeline is distinct from the genotype quality control described under Genome-Wide Association.

**Downstream evaluation in HPP**

Frozen representations were evaluated in the HPP cohort against eight systemic biomarkers — creatinine, hemoglobin, HDL cholesterol, AHI, heart rate, liver fat, glucose and white-blood-cell count — plus chronological age (continuous regression targets), and 37 self-reported chronic conditions (binary classification targets). To ensure adequate statistical power, stable model convergence, and valid FDR correction, binary conditions were restricted to chronic diagnoses with ≥100 positive cases (osteoporosis was excluded, as its diagnostic bone-mineral-density measure is itself a transformation of a DXA readout), following established minimum-case conventions for large-scale frameworks [64]. Performance was assessed under two paradigms: frozen linear probing and end-to-end fine-tuning. Both used repeated 80/20 participant-level train/validation splits across 10 random seeds with strictly disjoint subject sets between partitions.

**Linear probing.** Frozen embeddings were evaluated under a concatenation (early-fusion) strategy: bone- and tissue-modality embeddings were concatenated into a single fused representation. A single L2-penalised probe was fit on the fused representation — ridge regression for continuous targets, logistic regression for binary targets — with the penalty selected by nested 2-fold cross-validation (ridge penalty α over {0.1, 1, 10, 100, 1000}; logistic inverse-penalty C over $\{10^{-4}, 10^{-2}, 1, 10^{2}, 10^{4}\}$). For prevalent-condition classification in HPP, models were additionally compared under a four-arm covariate-adjusted design (covariates only; covariates +LeDXA; covariates + DINOv3; covariates + tabular DXA), all under concatenation, across 10 seeds with Wilcoxon signed-rank tests and Benjamini–Hochberg FDR correction over the 37 binary self-reported disease labels, so that reported gains reflect image-derived signal beyond age, sex, and BMI.

**End-to-end fine-tuning.** The LeDXA backbone and the DINOv3 baseline were fine-tuned under a unified protocol: AdamW with decoupled learning rates ($1\times10^{-5}$ backbone, $1\times10^{-4}$ heads), weight decay 0.01, 30 epochs with a OneCycle schedule [65] (cosine annealing, 5-epoch warmup), bfloat16 mixed precision, and gradient clipping (max norm 1.0). Two fusion strategies were compared: (i) late fusion, averaging per-modality head predictions (auxiliary loss weight 0.5), and (ii) concatenation fusion, concatenating bone and tissue representations before a single linear head. The best checkpoint was selected on validation performance.

**Disease grouping**. To evaluate model performance at the level of physiological systems rather than individual diagnoses, each tracked HPP diagnosis was assigned to one of 20 predefined organ-system categories based on standard clinical classification; a participant was labeled positive for a category if affected by any constituent diagnosis, regardless of whether that individual diagnosis met the ≥100-case threshold used for the single-disease analyses. Of these, 13 categories contained at least one constituent diagnosis and were retained for analysis; categories with no represented diagnosis, and the surgery-related category, were excluded from the grouped comparison. The full category-to-diagnosis mapping, including per-diagnosis case counts, is provided in Table S4.

**UKBB cross-sectional disease classification**

Prevalent-disease prediction in UKBB used a four-arm, covariate-adjusted design comparing: (1) covariates only (age, sex, BMI); (2) LeDXA embeddings + covariates; (3) DINOv3 Huge+ embeddings + covariates; and (4) DXA tabular features + covariates, all under early fusion. For each disease, a single shared 80/20 subject-level stratified split was applied across all arms, and an L2-penalised logistic-regression classifier was fit — inverse-penalty C selected by nested 2-fold cross-validation over the five-point log-spaced grid $C \in \{10^{-3}, 10^{-1.5}, 1, 10^{1.5}, 10^{3}\}$, lbfgs solver, ≤2,000 iterations — and scored by validation AUROC. The procedure was repeated over 10 random seeds. Consistent with the HPP analysis, diseases were restricted to chronic conditions with ≥100 positive cases in the shared analytic cohort (excluding osteoporosis); this resulted in 28 conditions being analyzed. Sex-specific conditions were restricted to the relevant sex. AUROC differences between models for each disease were tested with a two-tailed Wilcoxon signed-rank test (paired within seed) and corrected across diseases by the Benjamini–Hochberg FDR procedure.

**Longitudinal incident-disease risk**

Time-to-event analyses were anchored at the UKBB visit-2 DXA scan. Endpoints comprised incident first-occurrence diseases, defined from UK Biobank hospital-episode-statistics (HES)

"first reported" fields as the first ICD-10 diagnosis of each condition after baseline, together with all-cause mortality. Applying the same ≥100 minimum-incident threshold used in the cross-sectional analyses yielded 20 endpoints: type 2 diabetes (E11), knee arthrosis (M17), hip arthrosis (M16), other arthrosis (M19), osteoporosis (M81), chronic ischemic heart disease (I25), angina (I20), acute myocardial infarction (I21), heart failure (I50), atrial fibrillation (I48), cerebral infarction (I63), chronic renal failure (N18), acute renal failure (N17), chronic obstructive pulmonary disease (J44), liver disease (K76), hypothyroidism (E03), spondylopathies (M48), intervertebral disk disorders (M51); all-cause and cancer-specific mortality were evaluated as additional endpoints. Events were defined as the first hospital-episode-statistics (HES) record or death-registry entry occurring after baseline; prevalent cases (events on or before baseline) were excluded, and follow-up was right-censored at a fixed study-level administrative cutoff. Competing causes of death were treated as censoring events (cause-specific hazards).

The analysis cohort for each endpoint comprised participants with complete covariates (age, sex, and BMI), DXA-tabular features, and image embeddings (n = 22,481–25,945 across endpoints, depending on missingness and sex-specific eligibility). Separate Cox proportional-hazards models were fitted for LeDXA embeddings, DINOv3 embeddings, DXA-tabular features, and covariates alone. The embedding-based models used the first 100 principal components, which explained 90% and 87% of the cumulative variance in the LeDXA and DINOv3 representations, respectively; the DXA-tabular model used all 117 scanner-derived features without dimensionality reduction. All feature-based models were adjusted for age, sex, and BMI. Within each model, the covariate and feature blocks were assigned separate L2 penalties, selected for each endpoint from {0.01, 0.1, 1, 10, 100}. Discrimination was evaluated using a stratified 80/20 hold-out split repeated across 10 random seeds and quantified with Harrell's concordance index (C-index) [66]. Clinical utility was assessed as the cumulative proportion of incident events occurring among participants assigned to the highest predicted-risk quartile at baseline. Event capture was tracked over follow-up and compared between LeDXA and the DXA-tabular baseline at each time point (Fig. 3b), with differences assessed using paired bootstrapping.

## Biological age

**Gap derivation.** A ridge-regression model (L2 penalty selected by internal cross-validation) was trained to predict chronological age from the frozen LeDXA embeddings. Within each cohort, out-of-fold predicted ages were obtained by 10-fold subject-grouped (GroupKFold) cross-validation: for each fold the model was trained on the remaining folds and applied to the held-out subjects, so that no individual's predicted age depended on their own data. The biological-age gap was defined as the residual from an ordinary least-squares model regressing out-of-fold predicted age on a degree-2 polynomial of chronological age (age and age$^2$). This detrending removes both linear and quadratic age dependence, making the gap orthogonal to chronological age by construction. Empirically, the gap showed negligible correlation with chronological age overall ($r = -0.002$) and within each sex (females $r = +0.006$, males $r = -0.019$, Fig. S6). The gap showed a small mean offset by sex (females −0.13 yr, males +0.18 yr; point-biserial $r = +0.048$), a consequence of pooled detrending; all downstream phenotypic and prognostic analyses are conducted within sex. Within the UKBB cohort, subjects were ranked by gap and assigned to quartiles (Q1, biologically youngest, to Q4, biologically oldest) without sex stratification. Phenotypic profiles were compared between Q4 and Q1 using DXA-derived body-composition tabular features; effect-direction replication in HPP was assessed by Spearman correlation of standardized effect sizes for features available in both cohorts (Fig. S7).

**Prevalent disease stratification by age-gap quartile.** To characterize the cross-sectional disease burden associated with older-appearing DXA morphology, we evaluated a panel of 17 prevalent conditions at UKBB visit 2, spanning musculoskeletal (rheumatoid arthritis, osteoarthritis, osteoporosis, and back pain), cardiometabolic (diabetes, cardiovascular disease, atrial fibrillation, stroke, hypertension, obesity, and hyperlipidemia), respiratory (COPD, asthma, and sleep apnea), renal (renal failure), hepatic (liver disease), and oncologic (cancer) domains. Each condition was ascertained from ICD-10 hospital episode statistics (HES) where available and from self-reported diagnoses (UKBB Field 20002) otherwise; prevalent cases were defined as those recorded on or before the visit-2 date. Using the cohort-wide biological-age-gap quartiles defined above, prevalence was calculated separately in women and men. Q4-to-Q1 prevalence ratios (PRs) were estimated within each sex using Poisson regression with a log link and robust sandwich variance estimates. Monotonic gradients across ordered quartiles were assessed using the Cochran–Armitage trend test. Low muscle quantity was defined using the EWGSOP2 [45] ALM index thresholds of ALMI $<7.0$ kg/m$^2$ in men and $<5.5$ kg/m$^2$ in women. The association between each 1-year increase in the continuous biological-age gap and the odds of low ALMI was estimated separately by sex using logistic regression adjusted for chronological age. Q1-versus-Q4 differences in prevalence were tested with a two-sided $\chi^2$ test on the $2 \times 2$ count table; the Poisson models above provide the prevalence-ratio estimates. P values were corrected using the Benjamini–Hochberg procedure across the 54 sex-by-condition comparisons (27 conditions $\times$ 2 sexes: the 17 broad conditions plus the 10 ICD-10 sub-conditions obtained by decomposing cardiovascular disease, osteoarthritis and renal failure).

**All-cause mortality**. Prognostic analyses were conducted using the UKBB visit-2 cohort. After excluding participants whose recorded date of death preceded their baseline imaging assessment, the final analytic cohort comprised 39,310 individuals. Time to all-cause death was ascertained from death registry records, right-censored at the administrative cutoff (4 December 2022). Cox proportional-hazards regression was fit with the biological-age gap as the primary predictor, adjusted for chronological age and sex. Two parameterizations were evaluated: (i) a continuous model with the gap entered as a per-year increment, and (ii) a categorical model with quartile as an ordered factor (Q1 as reference). The proportional-hazards assumption was assessed using scaled Schoenfeld residuals. Kaplan–Meier survival curves were estimated for Q4 and Q1 separately.

**Medication response analysis.** To evaluate whether the biological-age gap responds to pharmacological treatment, HPP participants with at least two DXA scans were linked to longitudinal self-reported medication questionnaires. Each ATC level-3 drug class was assessed separately in women and men. Participants were included for a given class if their first self-reported treatment initiation fell strictly between their baseline and follow-up DXA scan dates, ensuring the pre-scan observation was treatment-naïve and the post-scan observation was post-treatment for that class. Twenty-three drug classes with sufficient participant counts were analyzed. For each included participant, the within-subject change in biological-age gap (Δgap = gap_follow-up − gap_baseline) was computed. Drug classes were tested using a one-sample Wilcoxon signed-rank test against a null of zero change. Testing was restricted to the 35 class-by-sex combinations retaining more than 10 paired participants, covering 22 of the 23 classes (beta-blocking agents, C07A, did not reach more than 10 paired participants in neither sex), Full results are summarized in Table S7.

**Statistical analysis**

**Evaluation metrics.** Continuous regression targets were scored by Pearson correlation (*r*) and binary classification targets by AUROC. All train/validation partitions were subject-disjoint, so that no individual appeared in both training and validation sets within a given fold or seed.

**Multiple-comparison correction.** P-values across multiple tests (e.g., across diseases, across biomarkers, across features) were corrected by the Benjamini–Hochberg false-discovery rate procedure [67]. For $m$ tests with p-values $p_1, p_2, \ldots, p_m$ sorted in ascending order, the critical threshold is

$$p_i \leq (i/m)\alpha, \text{ where } \alpha = 0.05$$

Adjusted p-values are computed and a test is considered significant if adjusted $P < 0.05$.

**Clustering**

LeDXA embeddings from the female HPP subset were standardized, projected onto 100 principal components, and embedded into two dimensions with UMAP [48]. K-means clustering (K = 2, 10 random initializations) was performed on the two-dimensional embedding, yielding a more adipose cluster (Cluster A) and a leaner cluster (Cluster B). DBSCAN (eps = 1.0, min_samples = 15) was then applied to the UMAP coordinates to remove outlier points in small isolated sub-clusters, retaining only the two largest density clusters. To isolate cluster-intrinsic biology from confounding by age and body size, a matched cohort was constructed by 1:1 Mahalanobis-distance [68] matching on standardized age and BMI. Cluster B cases were matched to Cluster A controls by greedy 1:1 nearest-neighbor matching without replacement, within a Mahalanobis caliper of 0.5, yielding 1,035 matched pairs; covariate balance was confirmed by standardized mean differences (SMD ≤ 0.03 for age and BMI) in the matched sample. Within the matched cohort, per-body-system volcano analyses used two-sided Mann-Whitney tests with Cohen's d as the effect size (mean difference divided by pooled standard deviation), with 95% confidence intervals. Multiple-testing correction used an alpha of 0.05.

**Multi-omics comparison of the embedding clusters.** The matched female cohort was compared across four molecular platforms: plasma proteomics (Olink), nuclear magnetic resonance metabolomics (Nightingale), gut metagenomics and whole-blood transcriptomics. Platform coverage of the matched cohort differs, so group sizes vary by platform: Olink 302 versus 321 participants across 1,026 proteins, NMR 748–775 versus 772–808 across 107 biomarkers (varying with per-biomarker missingness), metagenomics 998 versus 1,004 across 294 taxa, and transcriptomics 634 versus 617 across 1,000 transcripts (chosen by highest variability). Each feature was compared between clusters with a two-sided Mann-Whitney test, all significant and suggestive level hits are shown in Table S9.

**Genome-wide association of embedding space**

We assessed the genetic basis of the learned representations against a gold-standard benchmark. Three phenotype sets were built from visit-2 UK Biobank data for participants passing a DXA completeness filter. For the gold-standard set, each tabular DXA measurement was retained as an individual quantitative phenotype if at least 25,000 (~>50%) participants had a non-missing value, and no dimensionality reduction was applied. For the model set, raw DXA scans were passed through the trained LeDXA encoder under the same deterministic inference procedure used throughout (resized image, no augmentation); the embeddings of the two full body scans (bone and soft tissue) were concatenated column-wise, and the resulting high-dimensional space was reduced to its top 20 principal components, each component forming one phenotype. An identical pipeline was applied to DINOv3 embeddings of the same scans. Phenotypes were written keyed by sample identifier, with age and sex held in a parallel covariate table. Genotypes were taken from the merged UK Biobank array call set and passed through a single quality-control pipeline, applied independently to each cohort and restricted to its phenotyped samples.

Variants were filtered on minor allele frequency (at least 0.01), variant missingness (at most 0.02) and Hardy–Weinberg equilibrium ($P \geq 1\times10^{-6}$), after which samples exceeding 2% genotype missingness were removed. A linkage-disequilibrium-pruned variant set (50-variant window, 5-variant step, $r^2 < 0.2$) was used for the structure-sensitive steps that followed. Heterozygosity outliers more than three standard deviations from the mean rate were excluded, related individuals were removed using a KING kinship [69] cutoff of 0.0884 (pairs closer than third-degree relatives), and the first 20 genotype principal components were computed on the pruned, unrelated set; participants beyond four standard deviations on any of the first four components were dropped as ancestry outliers. A final minor allele frequency and missingness filter produced the analysis-ready genotype set.

Association testing was performed in PLINK 2 [70]. Each phenotype was regressed on variant dosage under an additive linear model, adjusting for age, sex and the first 20 genotype principal components, with covariates variance-standardized and their per-variant coefficients suppressed from the output. Genome-wide significance was defined at $P < 5\times10^{-8}$. Significant variants were aggregated across phenotypes within each representation by the minimum P value per variant, and lead SNPs were taken as the union of genome-wide significant variants. Independent genetic signals were defined separately within each representation (tabular, LeDXA, DINOv3) by LD clumping in PLINK (index $P < 5\times10^{-8}$, $r^2 < 0.1$, 500 kb window) against the in-sample quality-controlled UKBB genotypes, yielding 258 tabular, 23 DINOv3, and 73 LeDXA loci; loci were then compared for genomic-region overlap across representations to identify representation-specific signals, yielding 18 loci specific to the LeDXA embedding. Lead variants were mapped to genes by position using MyVariant.info [71] (dbSNP, SnpEff and CADD annotations). Overlap with established associations was assessed by querying the NHGRI-EBI GWAS Catalog [72] REST API for each lead SNP; where the exact variant returned no association, linkage disequilibrium ($r^2$ and D′, 1000 Genomes European reference) was assessed against every catalogued variant within the clumping window, since $r^2$ alone can be attenuated by allele-frequency differences between variants tagging the same haplotype. A locus was considered to tag a previously reported association if either its lead SNP or a variant in high LD/D′ had a catalogued association; documented associations were further classified by organ-system domain (e.g., body composition, bone, renal, pulmonary) for the LeDXA-specific loci (Fig. 4c). Per-phenotype SNP heritability was estimated by LD-score regression [73], munging the PLINK summary statistics and regressing them against precomputed LD scores from the LDSC reference panel to report observed-scale $h^2$ with standard errors.

## Data availability

Data in this paper is part of the Human Phenotype Project and is accessible to researchers from universities and other research institutions at: https://humanphenotypeproject.org/data-access

## Code availability

https://github.com/GilSasson1/LeDXA

## Acknowledgments

We thank the members of the Segal group for fruitful discussions. E.S. is supported by the Crown Human Genome Center; the Larson Charitable Foundation New Scientist Fund; the Else Kröner Fresenius Foundation; the White Rose International Foundation; the Ben B. and Joyce E. Eisenberg Foundation; the Nissenbaum Family; Marcos Pinheiro de Andrade and Vanessa Buchheim; Lady Michelle Michels; Aliza Moussaieff; and grants funded by the Minerva Foundation, with funding from the Federal German Ministry for Education and Research and by the European Research Council and the Israel Science Foundation. Funded by the European Union Project 101095540 IMMEDIATE. Views and opinions expressed are, however, those of the author(s) only and do not necessarily reflect those of the European Union. Neither the European Union nor the granting authority can be held responsible for them. The funders had no role in study design, data collection and analysis, decision to publish or preparation of the manuscript. G.S. thanks Shay Cohen Weinberger for her invaluable support and encouragement throughout this project.

## Additional information

Corresponding author: eran.segal@weizmann.ac.il

## Supplementary Appendix

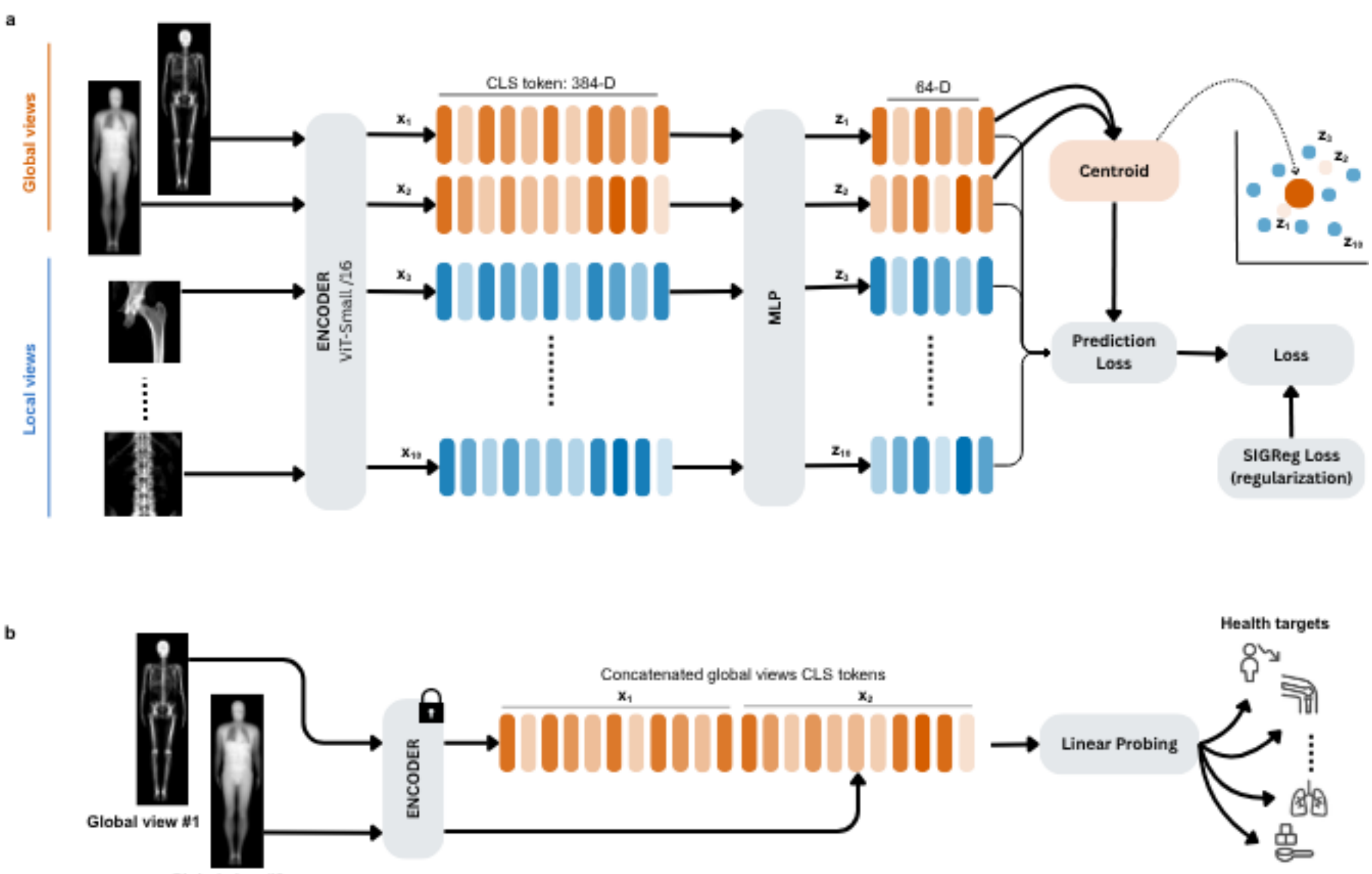

**Figure S1 | LeDXA architecture and evaluation setup. a,** Self-supervised pretraining with LeJEPA objective. Every DXA scan has two images: a bone image and a soft-tissue image, which are handled the same way. Each image is turned into 10 augmented views — 2 whole-body (global) views and 8 close-up (local) views (the femur/spine regional scans plus small random crops drawn from the whole body scan crops). A ViT-Small/16 shared encoder turns each view into a 384-dimensional representation (x1–x10), and a small 3-layer network compresses these to 64 dimensions (z1–z10). Training pushes every view's representation toward the average of the two global views, while a sketched isotropic gaussian regularization (SIGReg) term keeps the overall spread of representations close to a standard Gaussian distribution. **b,** The trained encoder is then frozen (its weights are not changed). For each scan, the bone and tissue representations are concatenated together and fed to a simple linear model that is trained to predict health outcomes.

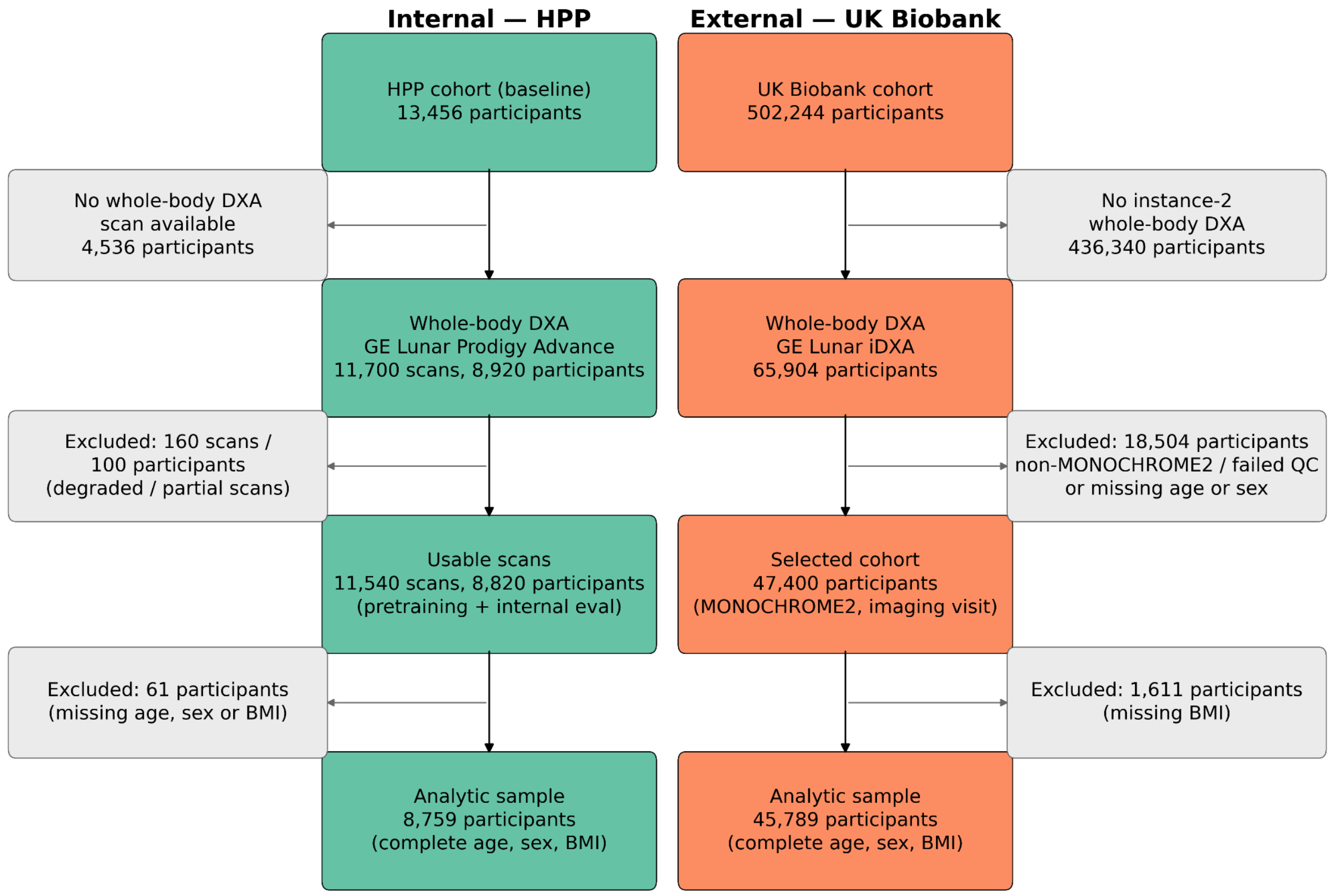

**Figure S2 | Study flow and cohort selection.** Participant flow for the internal Human Phenotype Project (HPP; green, left) and external UK Biobank (UKBB; orange, right) cohorts, detailing the progression from baseline enrollment to the final covariate-complete analytic samples. Colored boxes represent the cohorts retained at each step, while grey boxes indicate the number of excluded scans or participants alongside the reason for exclusion. For the HPP cohort, counts are provided for both dual-energy X-ray absorptiometry (DXA) scans and unique participants, as self-supervised pretraining utilized all available longitudinal scans, whereas the downstream analytic sample is defined at the participant level. UKBB counts are provided strictly per participant. Exclusions for both cohorts were limited to the availability of a usable whole-body DXA scan (format and image quality) and the completeness of baseline covariates (age, sex, and body mass index); no additional diagnosis- or medication-based exclusions were applied. QC, quality control.

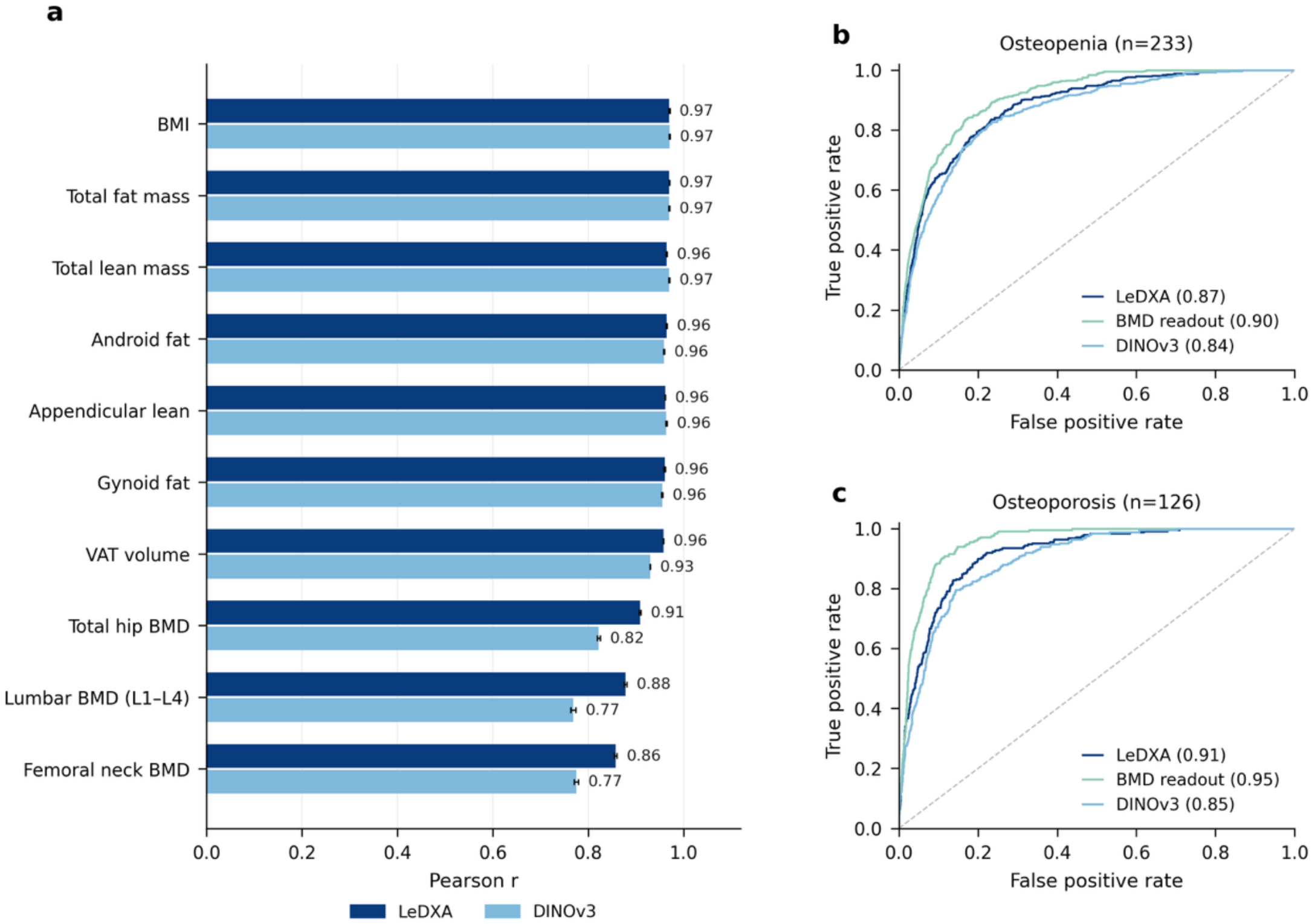

**Figure S3 | LeDXA recovers DXA-derived phenotypes from raw images (HPP). a,** Recovery of ten scanner-measured anthropometric and DXA quantities (body-mass-index, total fat mass, total and appendicular lean mass, android and gynoid fat, visceral adipose tissue (VAT) volume, and femoral-neck, total-hip and lumbar (L1–L4) bone-mineral density (BMD)) by the LeDXA and DINOv3 imaging embeddings (Pearson r between predicted and measured values; bars, mean over 10 random 80/20 subject-level splits; error bars, standard error of the mean). **b, c,** Receiver-operating-characteristic (ROC) curves for recovering self-reported osteopenia (b, n = 233) and osteoporosis (c, n = 126) from the LeDXA and DINOv3 embeddings, compared with a model trained on the scanner-derived BMD measurements. Legend values are the mean area under the ROC curve (AUROC); n, number of positive cases. Across the seven non-bone measures, both embeddings performed comparably (mean ± SD Pearson *r*: LeDXA 0.964 ± 0.005, DINOv3 0.957 ± 0.016), with LeDXA modestly ahead on VAT volume (0.96 vs. 0.93). On the three diagnosis-defining BMD sites, LeDXA scored higher than DINOv3 (0.882 ± 0.026 vs. 0.789 ± 0.029), mirroring its advantage over DINOv3 on the bone diagnoses in b, c.

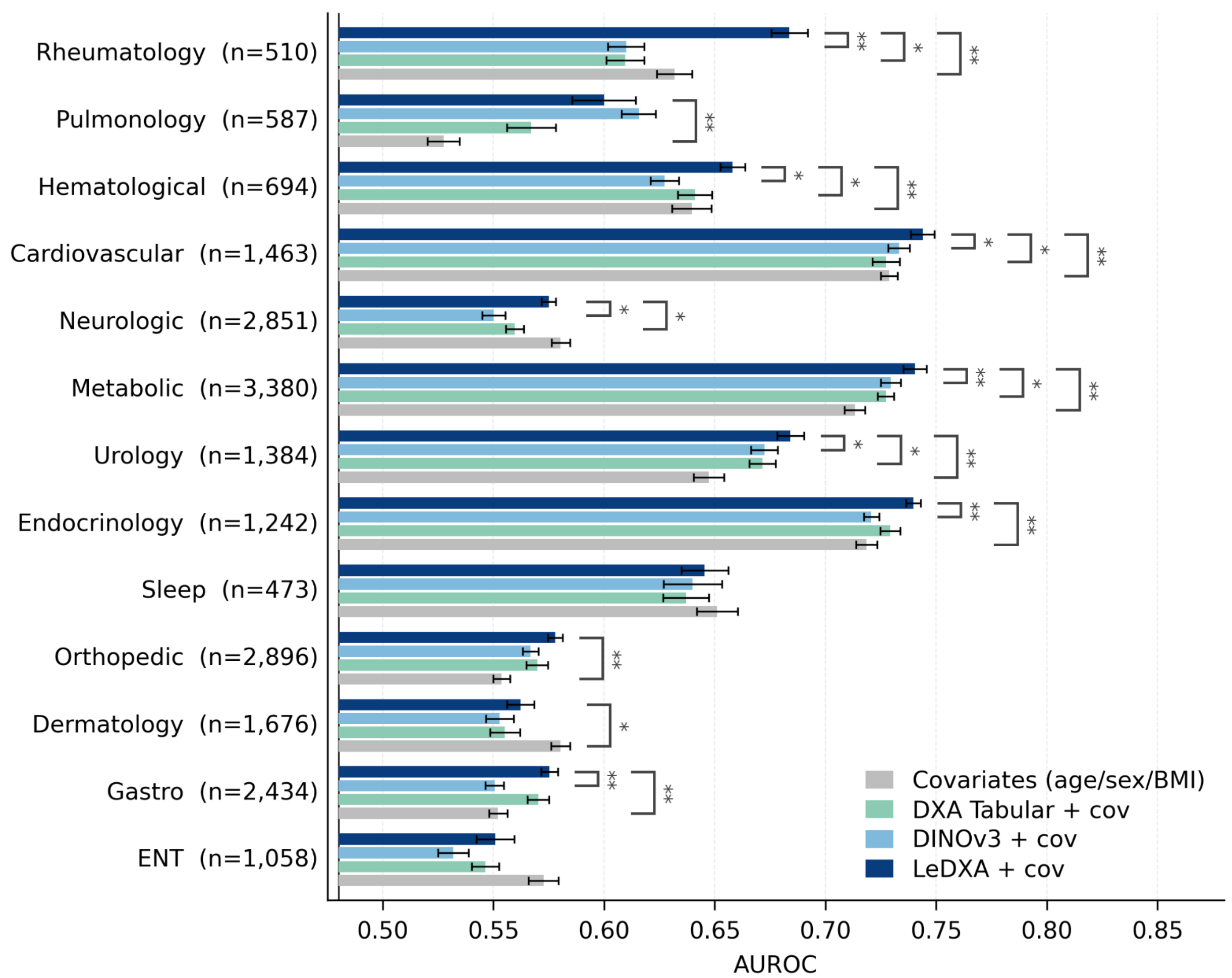

**Figure S4 | LeDXA's disease-predictive advantage across organ systems (HPP).** Mean AUROC for classifying organ-system-level disease endpoints in the HPP cohort. The performance of four models is compared: the LeDXA embedding, the general-purpose DINOv3 embedding, scanner-derived DXA-tabular features, and a covariate-only baseline (age, sex, and BMI). For the three main models, age, sex, and BMI were concatenated alongside the primary features (Methods). Individual diagnoses were grouped into 13 chronic organ-system categories (Methods; Table S4); a subject was labeled positive for a group if affected by any constituent diagnosis. Bars show the mean AUROC across 10 random 80/20 subject-level train/test splits, and error bars represent the standard error of the mean. The number of positive cases per group is indicated by n. Brackets denote two-sided Wilcoxon signed-rank tests across seeds, adjusted via the Benjamini–Hochberg FDR procedure across groups, comparing LeDXA with each of the other three arms (*adjusted $P < 0.05$, **adjusted $P < 0.01$). LeDXA significantly exceeds the DXA-tabular baseline in 6 of 13 groups, led by cardiovascular, hematological, metabolic, neurologic, rheumatological, and urological systems, indicating that the raw imaging representation carries systemic disease-predictive information not fully captured by predefined scalar DXA measurements.

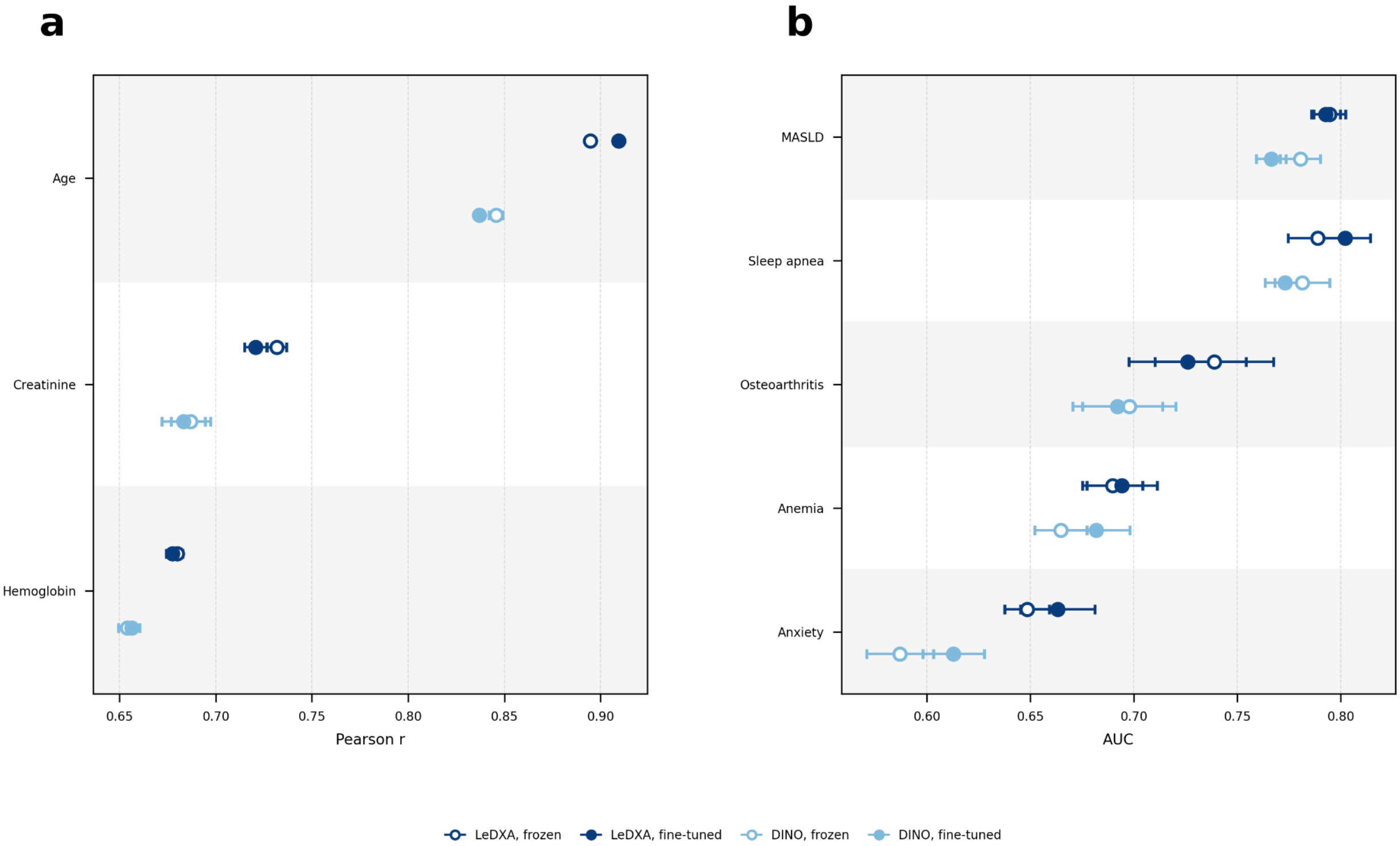

**Figure S5 | Supervised fine-tuning does not change model performance rankings.** Frozen linear probing (open markers) versus full end-to-end supervised fine-tuning (filled markers) for LeDXA and DINOv3, using embeddings alone (no age/sex/BMI covariates). Data represent mean ± standard error of the mean (s.e.m.) across 5 random seeds. **a,** Prediction of three continuous physiological biomarkers (chronological age, creatinine, and hemoglobin), evaluated by Pearson correlation (*r*). **b,** Classification performance (area under the receiver operating characteristic curve [AUROC]) for five prevalent conditions in the Human Phenotype Project (HPP) cohort: metabolic dysfunction-associated steatotic liver disease (MASLD), sleep apnea, osteoarthritis, anemia, and anxiety. Across all eight evaluated targets, supervised fine-tuning did not reverse the performance ranking of LeDXA relative to DINOv3 established under frozen probing.

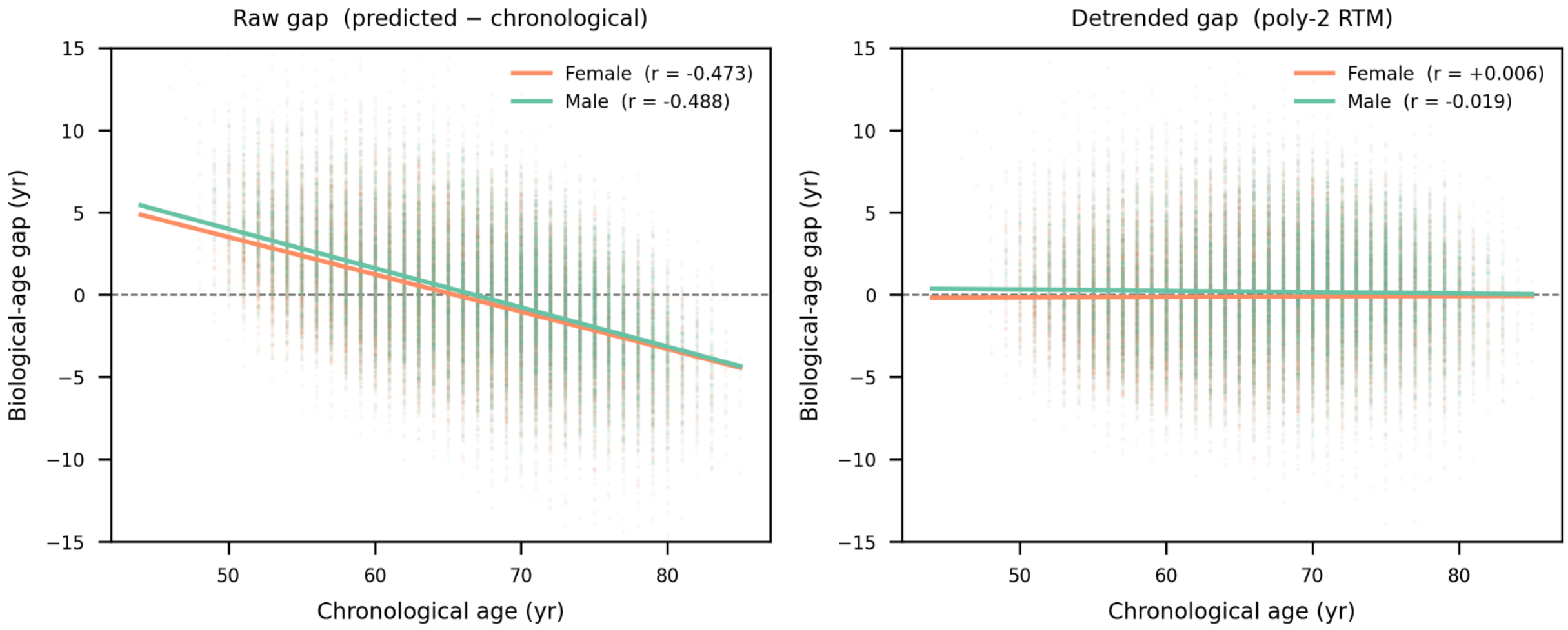

**Figure S6 | Biological-age gap versus chronological age, by sex.** Scatter of the detrended biological-age gap against chronological age in UK Biobank, showing negligible residual correlation within each sex.

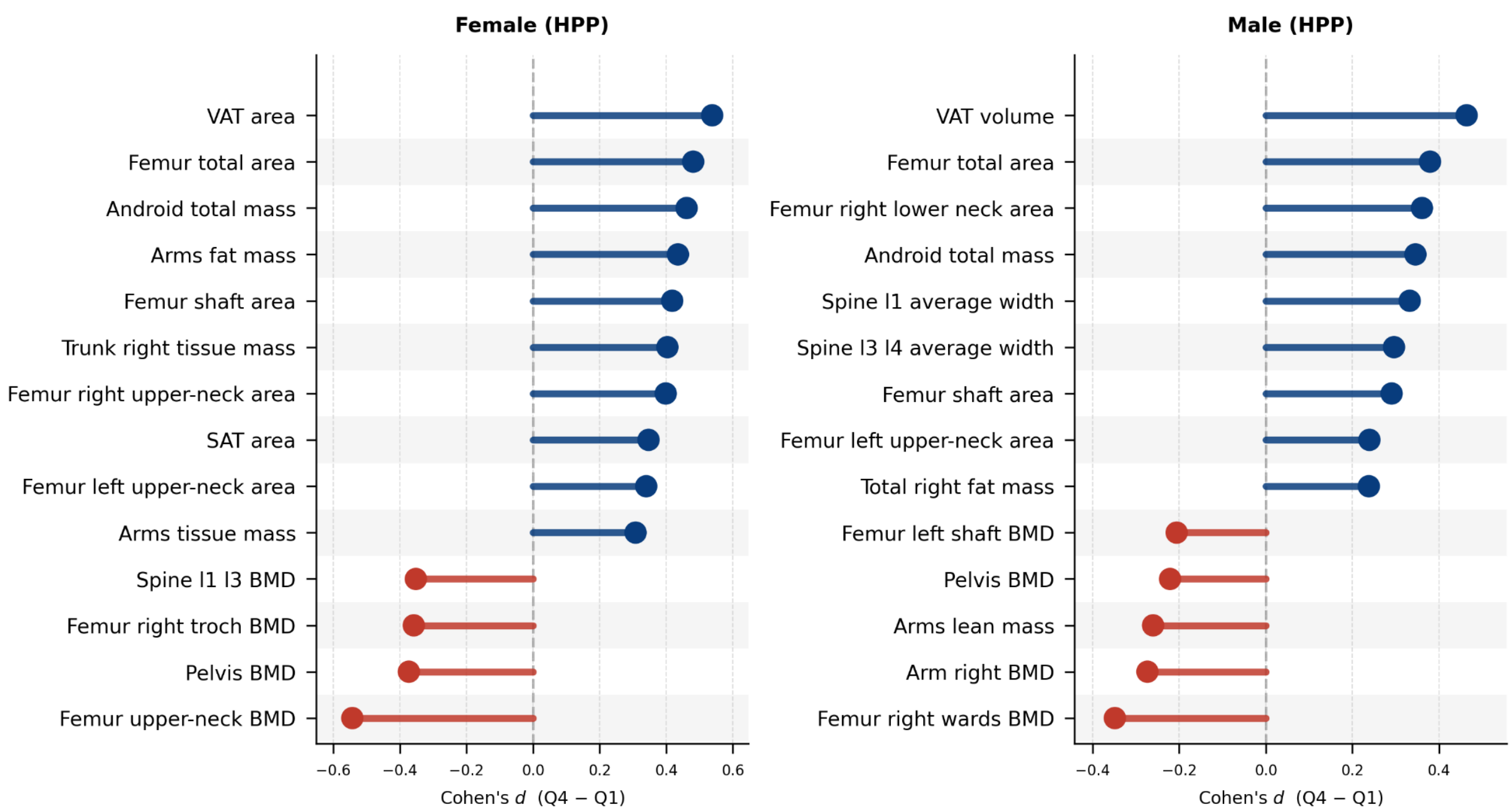

**Figure S7 | HPP replication of the biological-age-gap Q4-vs-Q1 DXA body-composition associations.** Cohen's d (Q4 − Q1) for the top false-discovery-rate-significant DXA body-composition and bone phenotypes in the HPP cohort, by sex, mirroring the UK Biobank analysis in Fig. 5c, d. Positive values indicate higher levels in the oldest-appearing (Q4) biological-age-gap quartile. Female n = 1,517 (Q1/Q4 = 380/380); male n = 1,221 (306/306). Higher visceral adipose tissue and lower femoral bone-mineral density in Q4 recapitulate the UK Biobank pattern.

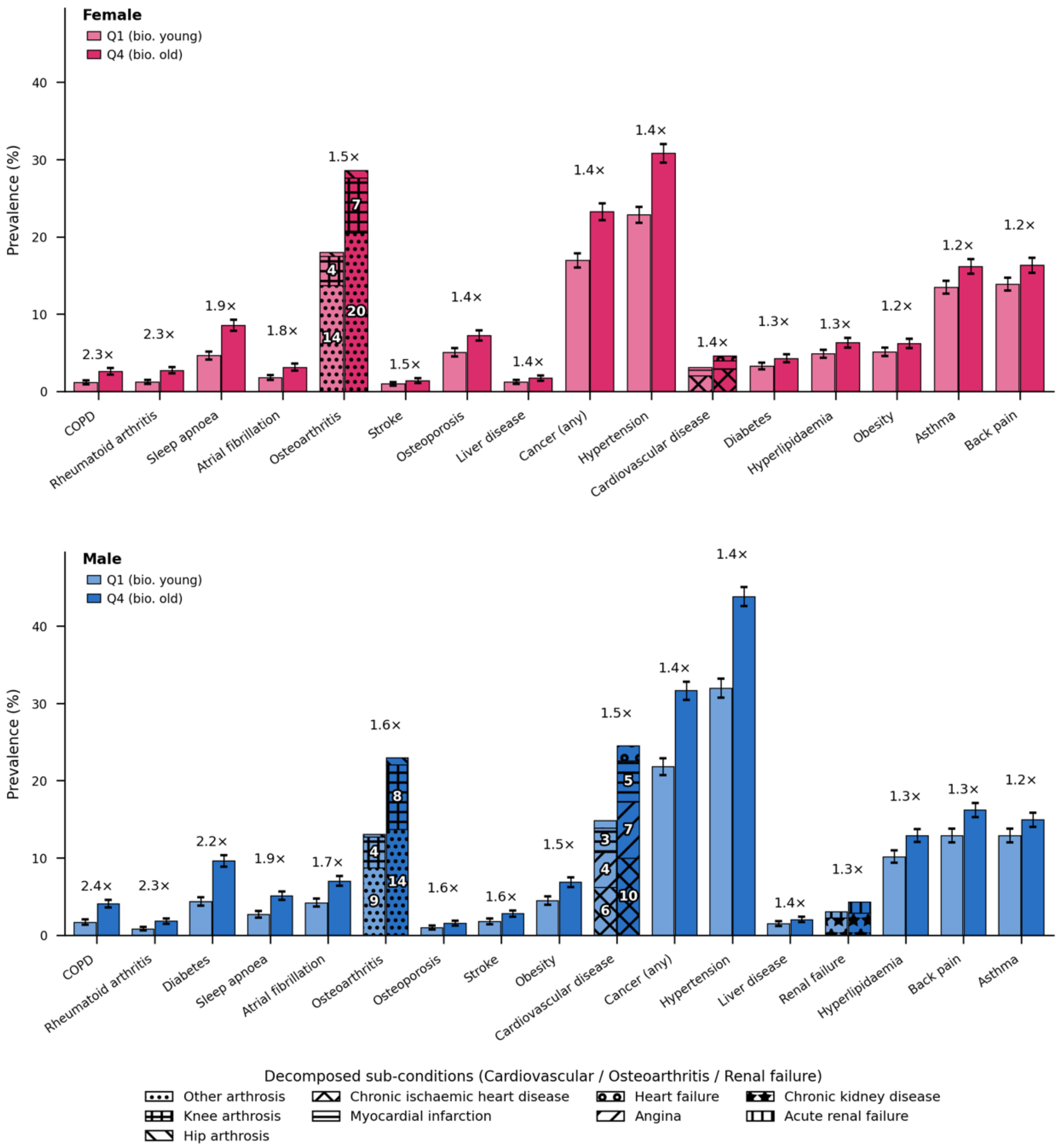

**Figure S8 | Cross-sectional disease prevalence by LeDXA biological-age-gap quartile in UK Biobank.** Prevalence of conditions present at the first imaging visit for participants in the lowest (Q1, biologically younger) versus highest (Q4, biologically older) quartile of the LeDXA biological-age gap, stratified by sex. The biological-age gap represents the residual of LeDXA-predicted age regressed on chronological age. A condition was considered prevalent if its earliest ICD-10/self-reported diagnosis preceded the imaging visit. Bars display percentage prevalence with 95% Wilson confidence intervals (error bars); values above each pair denote the Q4:Q1 prevalence ratio. Only conditions reaching significance (Benjamini–Hochberg false discovery rate (FDR) $< 0.05$, $\chi^2$ test) are shown; groups with fewer than 20 assessable participants were omitted.Three composite categories—cardiovascular disease, osteoarthritis, and renal failure—are represented as stacked bars decomposed into their constituent sub-conditions. Bar fill indicates quartile (light = Q1, dark = Q4), while hatch patterns denote specific sub-conditions. In-bar numbers report the sub-condition's prevalence (%). Because sub-conditions are not mutually exclusive (e.g., a participant may have both angina and myocardial infarction), the summed height of a decomposed bar can modestly exceed the true prevalence of its composite category.

**Table S1. Baseline characteristics of the study cohorts**

| Characteristic | Internal cohort (HPP) | External cohort (UKBB) |
|---|---|---|
| N | 8,759 | 45,789 |
| Country | Israel | United Kingdom |
| Age, years (mean ± SD, range) | 50.3 ± 9.7 (18–80) | 66.1 ± 7.6 (44–85) |
| Sex, women (n, %) | 4,558 (52.0%) | 23,729 (51.8%) |
| Sex, men (n, %) | 4,201 (48.0%) | 22,060 (48.2%) |
| BMI, kg/m² (mean ± SD, range) | 25.9 ± 4.1 (15.3–51.0) | 26.6 ± 4.5 (13.9–69.6) |

**Table S2. HPP prevalent-disease classification**

| Disease | Cases (n) | Covariates only | LeDXA + covariates | DINOv3 + covariates | DXA-tabular + covariates |
|---|---|---|---|---|---|
| Attention deficit hyperactivity disorder (ADHD) | 1,835 | 0.579 ± 0.005 | 0.592 ± 0.006 | 0.582 ± 0.005 | 0.579 ± 0.005 |
| Allergy | 2,246 | 0.590 ± 0.005 | 0.600 ± 0.005 | 0.598 ± 0.005 | 0.600 ± 0.005 |
| Anal fissure | 665 | 0.500 ± 0.010 | 0.523 ± 0.008 | 0.494 ± 0.011 | 0.508 ± 0.009 |
| Anemia | 523 | 0.667 ± 0.006 | 0.685 ± 0.007 | 0.674 ± 0.007 | 0.675 ± 0.006 |
| Anxiety | 390 | 0.627 ± 0.012 | 0.648 ± 0.014 | 0.639 ± 0.015 | 0.635 ± 0.014 |
| Asthma | 573 | 0.565 ± 0.012 | 0.569 ± 0.008 | 0.616 ± 0.009 | 0.575 ± 0.007 |
| Atopic dermatitis | 361 | 0.578 ± 0.010 | 0.572 ± 0.014 | 0.583 ± 0.013 | 0.569 ± 0.011 |
| Vitamin B12 deficiency | 528 | 0.565 ± 0.009 | 0.592 ± 0.007 | 0.581 ± 0.006 | 0.581 ± 0.009 |
| Back pain | 2,055 | 0.546 ± 0.004 | 0.579 ± 0.005 | 0.566 ± 0.003 | 0.574 ± 0.006 |
| Chronic sinusitis | 381 | 0.623 ± 0.009 | 0.617 ± 0.011 | 0.616 ± 0.011 | 0.614 ± 0.012 |

| Depression | 390 | 0.614 ± 0.011 | 0.632 ± 0.013 | 0.647 ± 0.010 | 0.618 ± 0.010 |
|---|---|---|---|---|---|
| Diabetes | 133 | 0.560 ± 0.022 | 0.738 ± 0.013 | 0.679 ± 0.019 | 0.668 ± 0.019 |
| Episodic vertigo | 276 | 0.616 ± 0.015 | 0.616 ± 0.015 | 0.632 ± 0.013 | 0.623 ± 0.014 |
| Fatty liver disease | 606 | 0.721 ± 0.008 | 0.774 ± 0.007 | 0.760 ± 0.007 | 0.745 ± 0.005 |
| Fibromyalgia | 129 | 0.725 ± 0.017 | 0.757 ± 0.017 | 0.747 ± 0.018 | 0.743 ± 0.015 |
| Folic acid deficiency | 100 | 0.607 ± 0.031 | 0.593 ± 0.022 | 0.543 ± 0.033 | 0.617 ± 0.024 |
| G6PD deficiency | 114 | 0.581 ± 0.012 | 0.572 ± 0.018 | 0.594 ± 0.018 | 0.524 ± 0.020 |
| Gallstone disease | 419 | 0.688 ± 0.014 | 0.714 ± 0.012 | 0.686 ± 0.013 | 0.692 ± 0.014 |
| Haemorrhoids | 2,167 | 0.522 ± 0.004 | 0.524 ± 0.004 | 0.521 ± 0.003 | 0.521 ± 0.004 |
| Headache | 254 | 0.628 ± 0.013 | 0.629 ± 0.015 | 0.631 ± 0.013 | 0.648 ± 0.018 |
| Hearing loss | 438 | 0.656 ± 0.014 | 0.655 ± 0.013 | 0.655 ± 0.013 | 0.659 ± 0.014 |
| Heart valve disease | 156 | 0.546 ± 0.015 | 0.529 ± 0.014 | 0.489 ± 0.015 | 0.514 ± 0.011 |
| Hyperlipidemia | 2,289 | 0.701 ± 0.005 | 0.735 ± 0.005 | 0.721 ± 0.005 | 0.727 ± 0.005 |
| Hypertension | 1,095 | 0.776 ± 0.007 | 0.801 ± 0.006 | 0.796 ± 0.006 | 0.787 ± 0.006 |
| Hyperthyroidism | 149 | 0.607 ± 0.026 | 0.622 ± 0.026 | 0.603 ± 0.024 | 0.603 ± 0.027 |
| Hypothyroidism | 696 | 0.670 ± 0.006 | 0.666 ± 0.007 | 0.659 ± 0.007 | 0.674 ± 0.006 |
| Irritable bowel syndrome (IBS) | 407 | 0.634 ± 0.007 | 0.643 ± 0.007 | 0.637 ± 0.007 | 0.643 ± 0.008 |
| Insomnia | 173 | 0.505 ± 0.014 | 0.522 ± 0.014 | 0.554 ± 0.019 | 0.558 ± 0.017 |

| Ischemic heart disease | 123 | 0.751 ± 0.013 | 0.752 ± 0.015 | 0.743 ± 0.010 | 0.756 ± 0.015 |
|---|---|---|---|---|---|
| Migraine | 623 | 0.658 ± 0.007 | 0.675 ± 0.006 | 0.661 ± 0.006 | 0.665 ± 0.008 |
| Osteoarthritis | 303 | 0.738 ± 0.013 | 0.760 ± 0.012 | 0.743 ± 0.011 | 0.734 ± 0.012 |
| Gastroduodenal ulcer | 207 | 0.614 ± 0.016 | 0.602 ± 0.018 | 0.602 ± 0.016 | 0.609 ± 0.015 |
| Peptic ulcer disease | 1,132 | 0.550 ± 0.011 | 0.561 ± 0.010 | 0.552 ± 0.007 | 0.553 ± 0.010 |
| Prediabetes | 832 | 0.680 ± 0.008 | 0.722 ± 0.007 | 0.712 ± 0.006 | 0.715 ± 0.007 |
| Psoriasis | 291 | 0.569 ± 0.015 | 0.530 ± 0.012 | 0.519 ± 0.011 | 0.546 ± 0.010 |
| Sleep apnea | 309 | 0.754 ± 0.011 | 0.774 ± 0.008 | 0.766 ± 0.009 | 0.753 ± 0.009 |
| Urinary tract stones | 307 | 0.698 ± 0.016 | 0.735 ± 0.013 | 0.712 ± 0.013 | 0.701 ± 0.014 |

**Table S3. UK Biobank prevalent-disease classification**

| Condition (ICD-10) | Cases (n) | Covariates only | LeDXA + covariates | DINOv3 + covariates | DXA-tabular + covariates |
|---|---|---|---|---|---|
| Anemia | 1,740 | 0.618 ± 0.004 | 0.651 ± 0.004 | 0.636 ± 0.005 | 0.634 ± 0.004 |
| Anxiety | 1,705 | 0.587 ± 0.006 | 0.622 ± 0.003 | 0.615 ± 0.004 | 0.596 ± 0.005 |
| Asthma | 3,544 | 0.551 ± 0.004 | 0.619 ± 0.004 | 0.619 ± 0.003 | 0.587 ± 0.003 |
| Atrial fibrillation | 942 | 0.713 ± 0.006 | 0.725 ± 0.005 | 0.717 ± 0.007 | 0.720 ± 0.006 |
| Back pain | 4,744 | 0.541 ± 0.003 | 0.588 ± 0.003 | 0.590 ± 0.004 | 0.592 ± 0.004 |
| Celiac | 401 | 0.568 ± 0.008 | 0.630 ± 0.005 | 0.659 ± 0.007 | 0.611 ± 0.005 |
| COPD | 500 | 0.658 ± 0.009 | 0.761 ± 0.007 | 0.765 ± 0.007 | 0.730 ± 0.007 |

| Depression | 2,818 | 0.611 ± 0.004 | 0.622 ± 0.003 | 0.624 ± 0.002 | 0.618 ± 0.003 |
|---|---|---|---|---|---|
| Diabetes | 1,138 | 0.726 ± 0.004 | 0.840 ± 0.003 | 0.836 ± 0.003 | 0.830 ± 0.003 |
| Endometriosis | 547 | 0.792 ± 0.003 | 0.802 ± 0.002 | 0.784 ± 0.003 | 0.802 ± 0.002 |
| Gallstone | 1,278 | 0.672 ± 0.005 | 0.683 ± 0.005 | 0.678 ± 0.005 | 0.682 ± 0.005 |
| Gout | 826 | 0.797 ± 0.005 | 0.816 ± 0.005 | 0.813 ± 0.005 | 0.810 ± 0.005 |
| Heart failure | 183 | 0.776 ± 0.011 | 0.783 ± 0.010 | 0.789 ± 0.010 | 0.786 ± 0.012 |
| Hyperlipidemia | 6,044 | 0.710 ± 0.002 | 0.736 ± 0.002 | 0.732 ± 0.002 | 0.735 ± 0.002 |
| Hypertension | 7,864 | 0.719 ± 0.002 | 0.743 ± 0.002 | 0.739 ± 0.002 | 0.741 ± 0.002 |
| Hyperthyroidism | 370 | 0.654 ± 0.007 | 0.665 ± 0.006 | 0.644 ± 0.009 | 0.652 ± 0.012 |
| Hypothyroidism | 1,682 | 0.685 ± 0.004 | 0.679 ± 0.004 | 0.674 ± 0.003 | 0.685 ± 0.003 |
| IBS | 1,930 | 0.592 ± 0.004 | 0.599 ± 0.003 | 0.585 ± 0.004 | 0.600 ± 0.005 |
| Ischemic heart disease | 1,652 | 0.733 ± 0.004 | 0.742 ± 0.004 | 0.734 ± 0.004 | 0.741 ± 0.004 |
| Kidney stones | 604 | 0.630 ± 0.008 | 0.655 ± 0.005 | 0.633 ± 0.008 | 0.653 ± 0.009 |
| Liver disease | 337 | 0.683 ± 0.012 | 0.693 ± 0.012 | 0.679 ± 0.011 | 0.686 ± 0.009 |
| Osteoarthritis | 4,328 | 0.647 ± 0.004 | 0.693 ± 0.005 | 0.696 ± 0.004 | 0.675 ± 0.004 |
| PCOS | 170 | 0.825 ± 0.007 | 0.831 ± 0.007 | 0.821 ± 0.006 | 0.827 ± 0.006 |
| Psoriasis | 889 | 0.546 ± 0.006 | 0.550 ± 0.009 | 0.545 ± 0.006 | 0.550 ± 0.007 |
| Renal failure | 699 | 0.687 ± 0.005 | 0.699 ± 0.003 | 0.689 ± 0.005 | 0.698 ± 0.004 |
| Rheumatoid arthritis | 392 | 0.594 ± 0.010 | 0.604 ± 0.008 | 0.580 ± 0.005 | 0.597 ± 0.011 |
| Sleep disorders | 779 | 0.619 ± 0.008 | 0.644 ± 0.007 | 0.629 ± 0.005 | 0.629 ± 0.006 |
| Stroke | 400 | 0.689 ± 0.008 | 0.688 ± 0.009 | 0.678 ± 0.007 | 0.692 ± 0.008 |

**Table S4. HPP disease organ-system grouping**

| Organ system | Diagnosis | Cases (n) | System total (n) |
|---|---|---|---|
| Cardiovascular | Hypertension | 1,095 | 1,463 |
| Cardiovascular | Heart valve disease | 156 | 1,463 |
| Cardiovascular | Ischemic heart disease | 123 | 1,463 |
| Cardiovascular | Erectile dysfunction | 56 | 1,463 |
| Cardiovascular | Loss of consciousness | 34 | 1,463 |
| Cardiovascular | Atrial fibrillation | 33 | 1,463 |
| Cardiovascular | Atherosclerotic disease | 32 | 1,463 |
| Cardiovascular | AV. conduction disorder | 21 | 1,463 |
| Cardiovascular | Myocarditis | 18 | 1,463 |
| Cardiovascular | Stroke | 7 | 1,463 |
| Cardiovascular | Heart failure | 6 | 1,463 |
| Cardiovascular | Pulmonary embolism | 5 | 1,463 |
| Cardiovascular | Aortic aneurysm | 4 | 1,463 |
| Dermatology | Allergy | 1,010 | 1,676 |
| Dermatology | Atopic dermatitis | 361 | 1,676 |
| Dermatology | Psoriasis | 291 | 1,676 |
| Dermatology | Vitiligo | 66 | 1,676 |
| Dermatology | Skin rash | 56 | 1,676 |
| Dermatology | Haemangioma | 12 | 1,676 |
| ENT | Hearing loss | 438 | 1,058 |
| ENT | Chronic sinusitis | 381 | 1,058 |
| ENT | Episodic vertigo | 276 | 1,058 |
| ENT | Tinnitus | 18 | 1,058 |
| ENT | Vertigo | 7 | 1,058 |
| Endocrinology | Hypothyroidism | 696 | 1,242 |
| Endocrinology | Osteopenia | 330 | 1,242 |
| Endocrinology | Osteoporosis | 174 | 1,242 |
| Endocrinology | Hyperthyroidism | 149 | 1,242 |
| Endocrinology | Hashimoto | 71 | 1,242 |
| Endocrinology | Goiter | 34 | 1,242 |
| Endocrinology | Hyperparathyroidism | 26 | 1,242 |
| Endocrinology | Vitamin d deficiency | 21 | 1,242 |
| Endocrinology | Pituitary gland adenoma | 6 | 1,242 |
| Endocrinology | Thyroid adenoma | 6 | 1,242 |
| Endocrinology | Parathyroid adenoma | 2 | 1,242 |
| Gastro | Peptic ulcer disease | 1,132 | 2,434 |
| Gastro | Vitamin B12 deficiency | 528 | 2,434 |
| Gastro | Gallstone disease | 419 | 2,434 |
| Gastro | Irritable bowel syndrome (IBS) | 407 | 2,434 |
| Gastro | Gastroduodenal ulcer | 207 | 2,434 |
| Gastro | Folic acid deficiency | 100 | 2,434 |
| Gastro | Lactose intolerance | 45 | 2,434 |
| Gastro | Celiac | 38 | 2,434 |
| Gastro | Inflammatory bowel disease (IBD) | 25 | 2,434 |
| Gastro | Iron deficiency | 23 | 2,434 |
| Gastro | FMF | 18 | 2,434 |
| Gastro | Gallbladder polyp | 11 | 2,434 |
| Gastro | Esophagitis | 10 | 2,434 |
| Gastro | Constipation | 5 | 2,434 |
| Gastro | Barrett's esophagus | 4 | 2,434 |
| Hematological | Anemia | 523 | 694 |
| Hematological | G6PD deficiency | 114 | 694 |

| Hematological | Thalassemia | 41 | 694 |
|---|---|---|---|
| Hematological | Hypercoagulability | 37 | 694 |
| Hematological | Lymphoma | 16 | 694 |
| Metabolic | Hyperlipidemia | 2,289 | 3,380 |
| Metabolic | Prediabetes | 832 | 3,380 |
| Metabolic | MASLD (Fatty liver disease) | 606 | 3,380 |
| Metabolic | Obesity | 562 | 3,380 |
| Metabolic | Diabetes | 133 | 3,380 |
| Metabolic | Primary hypercholesterolaemia | 79 | 3,380 |
| Metabolic | Overweight | 25 | 3,380 |
| Metabolic | Gestational diabetes | 10 | 3,380 |
| Neurologic | Attention deficit hyperactivity disorder (ADHD) | 1,835 | 2,851 |
| Neurologic | Migraine | 623 | 2,851 |
| Neurologic | Anxiety | 390 | 2,851 |
| Neurologic | Depression | 390 | 2,851 |
| Neurologic | Headache | 254 | 2,851 |
| Neurologic | Post trauma (PTSD) | 17 | 2,851 |
| Neurologic | Eating disorder | 10 | 2,851 |
| Neurologic | Bipolar disorder | 9 | 2,851 |

| Neurologic | Hydrocephalus | 5 | 2,851 |
|---|---|---|---|
| Neurologic | Neuropathy | 2 | 2,851 |
| Orthopedic | Back pain | 2,055 | 2,896 |
| Orthopedic | Fracture | 1,004 | 2,896 |
| Orthopedic | Intervertebral disc disease | 59 | 2,896 |
| Orthopedic | Meniscus tears | 49 | 2,896 |
| Orthopedic | PES PLANUS | 13 | 2,896 |
| Orthopedic | Tendon injury | 9 | 2,896 |
| Orthopedic | Intervertebral disc stenosis cervical region | 8 | 2,896 |
| Orthopedic | Tenosynovitis | 8 | 2,896 |
| Orthopedic | Tendon rupture | 4 | 2,896 |
| Pulmonology | Asthma | 573 | 587 |
| Pulmonology | COPD | 18 | 587 |
| Pulmonology | Pulmonary hypertension | 2 | 587 |
| Rheumatology | Osteoarthritis | 303 | 510 |
| Rheumatology | Fibromyalgia | 129 | 510 |
| Rheumatology | Gout | 98 | 510 |
| Sleep | Sleep apnea | 309 | 473 |
| Sleep | Insomnia | 173 | 473 |

| Urology | Urinary tract infection | 1,090 | 1,384 |
|---|---|---|---|
| Urology | Urinary tract stones | 307 | 1,384 |
| Urology | Renal stones | 51 | 1,384 |
| Urology | Urinary incontinence | 3 | 1,384 |

**Table S5. UK Biobank incident-disease Cox discrimination**

| Endpoint | Events (n) | Covariates only | LeDXA + covariates | DINOv3 + covariates | DXA-tabular + covariates |
|---|---|---|---|---|---|
| Osteoporosis | 159 | 0.755 ± 0.012 | 0.789 ± 0.011 | 0.781 ± 0.014 | 0.792 ± 0.011 |
| Hip Arthrosis | 288 | 0.620 ± 0.013 | 0.758 ± 0.008 | 0.673 ± 0.010 | 0.633 ± 0.014 |
| Knee Arthrosis | 285 | 0.671 ± 0.004 | 0.744 ± 0.006 | 0.716 ± 0.006 | 0.677 ± 0.005 |
| Spondylopathy | 113 | 0.694 ± 0.015 | 0.699 ± 0.015 | 0.700 ± 0.015 | 0.682 ± 0.019 |
| Arthrosis | 422 | 0.644 ± 0.012 | 0.659 ± 0.013 | 0.651 ± 0.014 | 0.650 ± 0.013 |
| Intervertebral Disk Disease | 134 | 0.590 ± 0.021 | 0.592 ± 0.020 | 0.599 ± 0.021 | 0.590 ± 0.022 |
| Type 2 Diabetes | 252 | 0.722 ± 0.008 | 0.753 ± 0.007 | 0.752 ± 0.010 | 0.721 ± 0.008 |
| Hypothyroidism | 117 | 0.640 ± 0.015 | 0.645 ± 0.015 | 0.641 ± 0.014 | 0.639 ± 0.015 |
| Heart Failure | 184 | 0.765 ± 0.008 | 0.766 ± 0.007 | 0.769 ± 0.008 | 0.763 ± 0.008 |
| Atrial Fibrillation | 400 | 0.735 ± 0.006 | 0.740 ± 0.006 | 0.741 ± 0.006 | 0.734 ± 0.007 |
| Ischaemic Heart Disease | 437 | 0.688 ± 0.006 | 0.691 ± 0.006 | 0.697 ± 0.005 | 0.687 ± 0.006 |
| Angina Pectoris | 209 | 0.679 ± 0.010 | 0.680 ± 0.010 | 0.680 ± 0.010 | 0.677 ± 0.010 |
| Myocardial Infarction | 167 | 0.659 ± 0.012 | 0.659 ± 0.012 | 0.660 ± 0.012 | 0.657 ± 0.012 |
| COPD | 101 | 0.678 ± 0.014 | 0.693 ± 0.011 | 0.684 ± 0.013 | 0.678 ± 0.014 |
| Chronic Renal Failure | 250 | 0.742 ± 0.007 | 0.744 ± 0.006 | 0.744 ± 0.006 | 0.742 ± 0.007 |
| Acute Renal Failure | 219 | 0.715 ± 0.013 | 0.718 ± 0.013 | 0.721 ± 0.013 | 0.715 ± 0.013 |
| Liver Disease | 206 | 0.671 ± 0.012 | 0.677 ± 0.012 | 0.671 ± 0.012 | 0.671 ± 0.012 |
| Cerebral Infarction | 119 | 0.655 ± 0.019 | 0.653 ± 0.019 | 0.655 ± 0.019 | 0.654 ± 0.019 |
| All-Cause Death | 275 | 0.715 ± 0.004 | 0.717 ± 0.004 | 0.734 ± 0.005 | 0.715 ± 0.004 |
| Cancer Death | 148 | 0.716 ± 0.012 | 0.717 ± 0.012 | 0.717 ± 0.011 | 0.716 ± 0.012 |

**Table S6. LeDXA embedding GWAS hits (genome-wide and suggestive; full table available in the released code; see Code availability)**

| Principal component | Genome-wide hits (n) | Suggestive hits (n) | Strongest hit | Locus (chr:position) | P | Nearest gene |
|---|---|---|---|---|---|---|
| PC11 | 117 | 75 | rs9594738 | 13:42,952,145 | $2.5\times10^{-35}$ | — |
| PC10 | 48 | 71 | rs72961013 | 6:127,529,780 | $5.5\times10^{-21}$ | uncharacterized LOC105377989 |
| PC12 | 41 | 71 | rs78110303 | 20:34,025,983 | $1.1\times10^{-18}$ | — |
| PC0 | 13 | 36 | rs1421085 | 16:53,800,954 | $6.2\times10^{-13}$ | FTO |
| PC14 | 10 | 63 | rs62027818 | 15:84,661,145 | $2.2\times10^{-10}$ | AC027807.1 |
| PC3 | 7 | 59 | rs3791675 | 2:56,111,309 | $1.5\times10^{-9}$ | EFEMP1 |
| PC6 | 7 | 18 | rs7979524 | 12:28,703,833 | $4.2\times10^{-11}$ | CCDC91 |
| PC8 | 5 | 24 | rs4711750 | 6:43,757,082 | $1.5\times10^{-14}$ | RNA polymerase I and III subunit C |
| PC1 | 2 | 30 | rs1562502 | 3:35,735,596 | $2.1\times10^{-9}$ | ARPP21 |
| PC18 | 2 | 18 | rs10235232 | 7:120,783,206 | $1.5\times10^{-8}$ | CPED1 |
| PC13 | 1 | 33 | rs12548347 | 8:72,530,375 | $2.8\times10^{-10}$ | — |
| PC5 | 1 | 11 | rs825475 | 12:124,567,308 | $3.1\times10^{-8}$ | FAM101A |
| PC15 | 0 | 30 | rs4931081 | 12:28,617,212 | $9.3\times10^{-8}$ | CCDC91 |
| PC16 | 0 | 21 | rs11715077 | 3:80,454,668 | $2.4\times10^{-6}$ | — |
| PC17 | 0 | 20 | rs71336078 | 3:136,489,602 | $8.2\times10^{-7}$ | — |
| PC19 | 0 | 15 | rs2236164 | 20:34,097,353 | $2.4\times10^{-7}$ | CEP250 |

| PC2 | 0 | 11 | rs772160 | 2:97,039,666 | $8.5\times10^{-7}$ | NCAPH |
|---|---|---|---|---|---|---|
| PC7 | 0 | 11 | rs67226117 | 15:39,671,752 | $3.5\times10^{-7}$ | uncharacterized LOC105370777 |
| PC4 | 0 | 9 | rs113299885 | 17:62,316,454 | $1.7\times10^{-6}$ | TEX2 |
| PC9 | 0 | 9 | rs2130410 | 6:33,962,404 | $1.4\times10^{-7}$ | — |

**Table S7. Paired biological-age-gap change by ATC level-3 medication class and sex**

| ATC-3 class | Drug class | Sex | Pairs (n) | Change in gap (yr) | Adj. P |
|---|---|---|---|---|---|
| G03F | Progestogens and estrogens in combination | Women | 11 | -2.66 | 0.034 |
| N06A | Antidepressants | Men | 17 | -2.961 | 0.037 |
| C09C | Angiotensin II receptor blockers (ARBs), plain | Men | 14 | -2.555 | 0.061 |
| B03B | Vitamin B12 and folic acid | Women | 52 | -1.1 | 0.087 |
| B03A | Iron antianemic preparations | Women | 27 | -1.077 | 0.287 |
| N05C | Hypnotics and sedatives | Women | 11 | -1.983 | 0.313 |
| A11C | Vitamin A and D, incl. combinations of the two | Women | 117 | -0.489 | 0.765 |
| N06A | Antidepressants | Women | 21 | -1.065 | 0.888 |
| A10B | Blood glucose lowering drugs, excl. insulins | Men | 35 | 0.707 | 0.890 |

| C10A | Lipid modifying agents, plain | Men | 89 | 0.225 | 0.890 |
|---|---|---|---|---|---|
| B01A | Antithrombotic agents | Men | 20 | -1.022 | 0.890 |
| A12A | Calcium supplements | Men | 22 | -0.734 | 0.890 |
| A11G | Ascorbic acid (vitamin C), incl. combinations | Women | 29 | 0.646 | 0.890 |
| M05B | Drugs affecting bone structure and mineralization | Women | 16 | -0.338 | 0.890 |
| G03D | Progestogen sex hormones and modulators of the genital system | Women | 12 | 0.463 | 0.890 |
| A02B | Drugs for peptic ulcer and gastro-oesophageal reflux disease (GORD) | Women | 14 | 0.475 | 0.890 |
| A11C | Vitamin A and D, incl. combinations of the two | Men | 73 | -0.233 | 0.890 |
| C09D | Angiotensin II receptor blockers (ARBs), combinations | Women | 11 | -0.549 | 0.890 |
| C10B | Lipid modifying agents, combinations | Men | 14 | 0.293 | 0.890 |
| C10A | Lipid modifying agents, plain | Women | 86 | 0.132 | 0.890 |
| A12C | Other mineral supplements | Women | 114 | -0.23 | 0.890 |
| C09D | Angiotensin II receptor blockers (ARBs), combinations | Men | 14 | -0.261 | 0.890 |
| G04C | Drugs used in benign prostatic hypertrophy | Men | 15 | 0.072 | 0.890 |
| B03B | Vitamin B12 and folic acid | Men | 51 | -0.151 | 0.890 |
| A10B | Blood glucose lowering drugs, excl. insulins | Women | 52 | 0.317 | 0.890 |
| A11D | Vitamin B1, plain and in combination with vitamin B6 and B12 | Women | 15 | -0.753 | 0.914 |
| A12A | Calcium supplements | Women | 51 | -0.371 | 0.923 |
| G03C | Estrogens, sex hormones and modulators of the genital system | Women | 44 | -0.194 | 0.923 |
| A12C | Other mineral supplements | Men | 74 | -0.092 | 0.923 |
| C09C | Angiotensin II receptor blockers (ARBs), plain | Women | 12 | -0.332 | 0.923 |
| A11G | Ascorbic acid (vitamin C), incl. combinations | Men | 20 | 0.067 | 0.982 |
| A02B | Drugs for peptic ulcer and gastro-oesophageal reflux disease (GORD) | Men | 12 | -0.263 | 1.000 |
| G02C | Other gynecologicals | Women | 11 | 0.125 | 1.000 |
| C10B | Lipid modifying agents, combinations | Women | 13 | 0.158 | 1.000 |
| B03A | Iron antianemic preparations | Men | 11 | 0.084 | 1.000 |

**Table S8. Female embedding-cluster matched phenotypic differences (full table available in the released code; see Code availability)**

| System | Significant in system (n) | Feature | Cluster A / B mean | Cohen's d (B − A), 95% CI | Adj. P |
|---|---|---|---|---|---|
| Bone density | 111 | Body pelvis area | 265 / 275 cm² | 0.360 (0.271 to 0.448) | $7.0\times10^{-12}$ |
| Bone density | 111 | Body pelvis BMC | 251 / 267 g | 0.335 (0.246 to 0.423) | $6.0\times10^{-10}$ |
| Bone density | 111 | Femur right troch BMC | 7.92 / 8.49 g | 0.282 (0.193 to 0.372) | $8.9\times10^{-10}$ |
| Body composition | 81 | Legs fat mass | 8.56 / 6.97 kg | -0.942 (-1.034 to -0.850) | $5.4\times10^{-90}$ |
| Body composition | 81 | Leg left fat mass | 4.25 / 3.46 kg | -0.936 (-1.028 to -0.844) | $3.4\times10^{-89}$ |
| Body composition | 81 | Leg right fat mass | 4.31 / 3.51 kg | -0.933 (-1.025 to -0.840) | $9.1\times10^{-89}$ |
| Cardiovascular | 17 | Resting heart rate (bpm) | 61.2 / 58.5 bpm | -0.322 (-0.409 to -0.234) | $2.3\times10^{-11}$ |
| Cardiovascular | 17 | RR interval (ms) | 999 / 1050 ms | 0.336 (0.248 to 0.423) | $2.3\times10^{-11}$ |
| Cardiovascular | 17 | QT interval (ms) | 421 / 431 ms | 0.334 (0.246 to 0.421) | $2.9\times10^{-11}$ |
| Sleep | 13 | Heart rate mean during wake | 66 / 63.5 bpm | -0.333 (-0.433 to -0.232) | $8.8\times10^{-8}$ |
| Sleep | 13 | Heart rate mean during REM | 64.2 / 61.8 bpm | -0.319 (-0.420 to -0.218) | $8.8\times10^{-8}$ |
| Sleep | 13 | Heart rate mean during sleep | 62.3 / 59.9 bpm | -0.321 (-0.422 to -0.221) | $8.8\times10^{-8}$ |
| Diet (nutrient intake) | 8 | Energy / BMR | 1.12 / 1.17 ratio | 0.202 (0.111 to 0.294) | 0.002 |
| Diet (nutrient intake) | 8 | Mediterranean diet score (IMEDAS) | 9.99 / 10.3 ordinal (questionnaire) | 0.192 (0.100 to 0.283) | 0.003 |
| Diet (nutrient intake) | 8 | Vitamin B-6 | 0.00101 / 0.00105 | 0.178 (0.087 to 0.270) | 0.005 |
| Frailty | 6 | Arm right lean mass | 1.89 / 2.07 kg | 0.560 (0.471 to 0.650) | $1.9\times10^{-32}$ |
| Frailty | 6 | Arm left lean mass | 1.8 / 1.96 kg | 0.517 (0.428 to 0.606) | $5.4\times10^{-29}$ |
| Frailty | 6 | Hand grip left | 51.5 / 56.6 lb | 0.453 (0.366 to 0.541) | $6.2\times10^{-23}$ |
| Hematopoietic | 5 | Mean Corpuscular Hemoglobin (MCH) | 30.4 / 30.8 pg | 0.222 (0.134 to 0.310) | $1.9\times10^{-6}$ |
| Hematopoietic | 5 | Mean Cellular Volume (MCV) | 89.5 / 90.2 fL | 0.162 (0.074 to 0.250) | $3.8\times10^{-4}$ |
| Hematopoietic | 5 | Red Cell Distribution Width (RDW) | 13.4 / 13.2 % | -0.166 (-0.254 to -0.077) | $3.8\times10^{-4}$ |
| Lifestyle | 5 | Physical activity moderate days a week | 2.35 / 2.95 days/week | 0.288 (0.182 to 0.394) | $1.2\times10^{-6}$ |
| Lifestyle | 5 | Vigorous activity minutes | 17.7 / 24.9 min/day | 0.236 (0.123 to 0.348) | $1.8\times10^{-5}$ |
| Lifestyle | 5 | Walking minutes day | 34.9 / 42.4 min/day | 0.215 (0.108 to 0.321) | 0.003 |
| Diet (questionnaire) | 3 | Water glasses day | 6.43 / 7.48 glasses/day | 0.262 (0.139 to 0.385) | $1.5\times10^{-4}$ |
| Diet (questionnaire) | 3 | Raw veg tablespoons day | 8.43 / 9.64 tablespoons/day | 0.156 (0.030 to 0.282) | 0.012 |
| Diet (questionnaire) | 3 | Cooked veg tablespoons day | 4.08 / 4.95 tablespoons/day | 0.191 (0.044 to 0.339) | 0.013 |
| Mental health | 3 | Health satisfaction | 4.34 / 4.55 ordinal (questionnaire) | 0.256 (0.161 to 0.352) | $6.1\times10^{-6}$ |
| Mental health | 3 | Tired or little energy fortnight | 1.15 / 0.987 ordinal (questionnaire) | -0.200 (-0.296 to -0.105) | 0.001 |
| Mental health | 3 | Happiness level | 4.26 / 4.38 ordinal (questionnaire) | 0.169 (0.073 to 0.265) | 0.007 |
| Family history | 1 | Age 10 weight | 2.01 / 2.18 ordinal (questionnaire) | 0.180 (0.064 to 0.296) | 0.019 |
| Liver | 1 | Liver sound speed | 1560 / 1570 m/s | 0.235 (0.147 to 0.323) | $2.0\times10^{-6}$ |

| Renal function | 1 | Creatinine | 0.727 / 0.758 mg/dL | 0.294 (0.138 to 0.451) | $3.2\times10^{-4}$ |
|---|---|---|---|---|---|

**Table S9. Female embedding-cluster multi-omics**

| Omics layer (scale) | Feature | Mean difference (B − A) | Cluster A / B (n) | Adj. P |
|---|---|---|---|---|
| Metabolites (NMR) — Platform concentration units | Creatinine | 2.31 | 773 / 808 | $9.7\times10^{-6}$ |
| Metabolites (NMR) — Platform concentration units | GlycA | -0.0222 | 775 / 808 | 0.004 |
| Microbiome — Relative abundance | Oscillibacter sp900544615 | 0.000108 | 998 / 1,004 | 0.078 |
| Microbiome — Relative abundance | Ruthenibacterium lactatiformans | -0.000305 | 998 / 1,004 | 0.078 |
| Microbiome — Relative abundance | Unclassified taxon | 2.72e-05 | 998 / 1,004 | 0.078 |
| Proteomics (Olink) — $\log_2$ scale | MYL3 | 0.472 | 302 / 321 | $9.3\times10^{-6}$ |
| Proteomics (Olink) — $\log_2$ scale | LRIG1 | -0.147 | 302 / 321 | 0.003 |
| Proteomics (Olink) — $\log_2$ scale | ITGB6 | 0.124 | 302 / 321 | 0.013 |
| Proteomics (Olink) — $\log_2$ scale | EXOSC3 | -0.297 | 302 / 321 | 0.052 |

**Table S10. DXA tabular feature dictionary (HPP and UK Biobank; full table available in the released code; see Code availability)**

| Measurement category | HPP (n) | UK Biobank (n) | Total (n) |
|---|---|---|---|
| Body composition | 132 | 47 | 179 |
| Bone mineral density (BMD) | 54 | 43 | 97 |
| Bone area and geometry | 76 | 14 | 90 |
| Bone mineral content (BMC) | 54 | 13 | 67 |
| Scan acquisition mode | 2 | — | 2 |
| All categories | 318 | 117 | 435 |